\documentclass[preprint,times,sort&compress]{elsarticle}
\usepackage{framed}
\usepackage{multirow}
\usepackage{amssymb}
\usepackage{latexsym}
\usepackage{url}
\usepackage{xcolor}
\usepackage{hyperref}
\usepackage{subcaption}
\usepackage{amsmath}
\usepackage{listings}
\usepackage{booktabs}
\usepackage{amsthm}
\usepackage{pifont}
\usepackage{graphicx}

\graphicspath{{images/}}

\newcommand{\norm}[1]{\left\lVert#1\right\rVert}
\newcommand{\codeword}[1]{\texttt{\textcolor{orange}{#1}}}
 \newcommand{\xmark}{\ding{55}}

\definecolor{newcolor}{rgb}{.8,.349,.1}
\definecolor{codegreen}{rgb}{0.7,0.7,0.7}
\definecolor{codegray}{rgb}{0.5,0.5,0.5}
\definecolor{codepurple}{rgb}{0.58,0,0.82}
\definecolor{backcolour}{rgb}{1,1,1}
\lstdefinestyle{mystyle}{ backgroundcolor=\color{backcolour},   
commentstyle=\color{codegreen}, keywordstyle=\color{magenta},
numberstyle=\tiny\color{codegray}, stringstyle=\color{codepurple},
basicstyle=\ttfamily\footnotesize, breakatwhitespace=false,         
breaklines=true,                 
captionpos=b,                    
keepspaces=true,                 
numbersep=5pt,                  
showspaces=false,                
showstringspaces=false, showtabs=false,                  
tabsize=2 }
\title{Altering Backward Pass Gradients improves Convergence}

\author[1]{Bishshoy Das\corref{cor1}}
\ead{bishshoy.das@ee.iitd.ac.in}

\author[1]{Milton Mondal\corref{cor2}}
\ead{milton.mondal@ee.iitd.ac.in}

\author[1]{\\ Brejesh Lall}
\ead{brejesh@ee.iitd.ac.in}

\author[1]{Shiv Dutt Joshi}
\ead{sdjoshi@ee.iitd.ac.in}

\author[1]{Sumantra Dutta Roy}
\ead{sumantra@ee.iitd.ac.in}

\cortext[cor1]{Corresponding author}
\cortext[cor2]{Equal contribution}

\address[1]{Electrical Engineering Department, Indian Institute of Technology Delhi\\ Hauz Khas, New Delhi - 110016, India}

\begin{document}

\begin{abstract}
In standard neural network training, the gradients in the backward pass are determined
by the forward pass. As a result, the two stages are coupled. This is how most neural
networks are trained currently. However, gradient modification in the backward pass has
seldom been studied in the literature. In this paper we explore decoupled training,
where we alter the gradients in the backward pass. We propose a simple yet powerful
method called PowerGrad Transform, that alters the gradients before the weight update in
the backward pass and significantly enhances the predictive performance of the neural
network. PowerGrad Transform trains the network to arrive at a better optima at
convergence. It is computationally extremely efficient, virtually adding no additional
cost to either memory or compute, but results in improved final accuracies on both the
training and test sets. PowerGrad Transform is easy to integrate into existing training
routines, requiring just a few lines of code. PowerGrad Transform accelerates training
and makes it possible for the network to better fit the training data. With decoupled
training, PowerGrad Transform improves baseline accuracies for ResNet-50 by 0.73\%, for
SE-ResNet-50 by 0.66\% and by more than 1.0\% for the non-normalized ResNet-18 network
on the ImageNet classification task.
\end{abstract}


\maketitle

\section{Introduction}
\label{sec:Intr}

\begin{figure}[t]
\centering
\includegraphics[width=\textwidth]{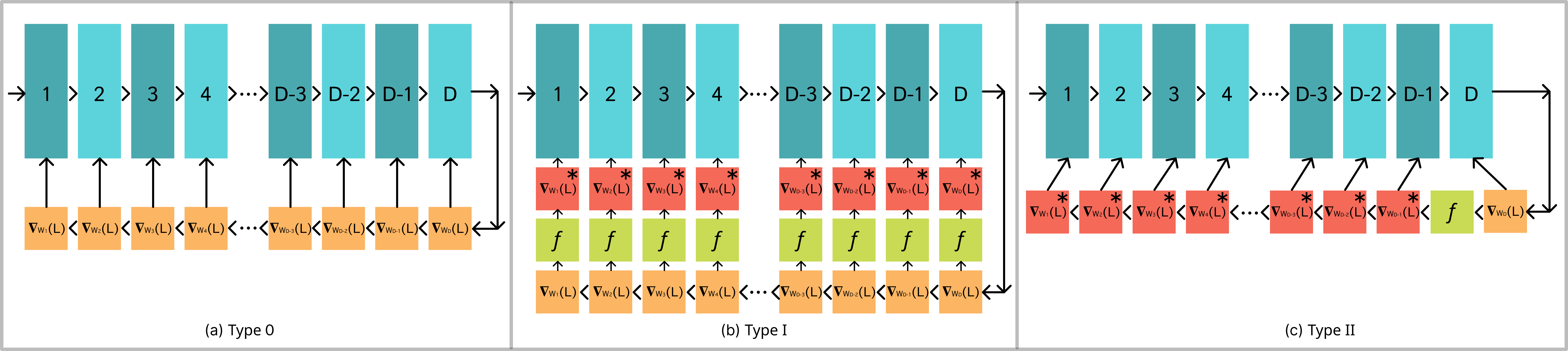}
\caption{ Different ways of altering gradients in the backward pass. Blue blocks denote
a CNN's different layers (convolutional, linear, batchnorm, softmax, etc.). Orange
blocks indicate the backward graph with unmodified gradients. Green blocks represent
transformation functions, while red blocks indicate transformed gradients. (a)
\textbf{Type 0:} Backpropagation without any gradient modification. (b) \textbf{Type I:}
Altering gradients after they are all computed. (c) \textbf{Type II:} Altering gradients
at an earlier stage in the backward computation graph, subsequently leading to
alteration of all other gradients. }
\label{fig:gradient_altering}
\end{figure}

Backpropagation is traditionally used to train deep neural networks
\cite{lillicrap2020backpropagation}. Gradients are computed using basic calculus
principles to adjust the weights during backpropagation \cite{lecun1988theoretical}.
Alternatives to traditional gradients has rarely been studied in the literature
hitherto. In normal training procedures, gradients are computed immediately in the
backward pass utilizing the values obtained in the forward pass. This makes the backward
pass coupled with the forward pass. However, decoupling the backward pass from the
forward pass by modifying the gradients to improve training efficiency and final
convergent accuracy has hardly been explored. In this paper we explore the landscape of
decoupling the forward pass from the backward pass by altering the gradients and
subsequently updating the network's parameters with the modified gradients. There are
several ways to achieve gradient modification in the backward pass. We discuss a few
techniques in Fig. \ref{fig:gradient_altering}. One could modify the gradients at either
\textbf{(I)} multiple points in the backward computation graph or \textbf{(II)} at a
point very early in the backward graph and allow this changed gradient to also affect
the rest of the network's gradients as it backpropagates through the backward
computation graph.

\textbf{Type 0:} No modification: In this method, we calculate the gradients using the
standard calculus rules and use the chain rule to calculate the gradients of the rest of
the network's parameters, also known as backpropagation as portrayed in Fig.
\ref{fig:gradient_altering}(a). The network is then updated with the gradient descent
equation:

\begin{equation}
W_i^{t+1} = W_i^t - \lambda\ \nabla_{W_i}(L)
\hspace{2cm}
i=D,D-1,\ldots,1
\end{equation}

\textbf{Type I:} Gradient modification at multiple points independently: In this method,
the gradients are first computed using standard procedure and then individually altered
as shown in Fig. \ref{fig:gradient_altering}(b). Gradient clipping
\cite{pascanu2013difficulty} and Adaptive gradient clipping \cite{brock2021high} are
examples of such modifications. In both of these methods, the gradients are first
computed using standard rules and then they are modified using some function. It can be
described as:

\begin{equation}
W_i^{t+1} = W_i^t - \lambda\ f(\nabla_{W_i}(L))
\hspace{2cm}
i=D,D-1,\ldots,1
\end{equation}

where the gradients $\nabla_{W_i}(L)$ are transformed using the transformation function
`$f$' before the weight update.

\textbf{Type II:} Gradient modification at a point very early in the backward graph: In
this type of modification, the gradient is altered at a very early stage in the backward
computation graph and then all subsequent gradients are generated using the values
obtained with the modified gradients. We illustrate this type of modification in Fig.
\ref{fig:gradient_altering}(c). Because of the chain rule, network parameters whose
gradients are connected to the point of alteration in the computation graph also gets
subsequently altered. It can be described as:

\begin{align}
W_D^{t+1} &= W_D^t - \lambda\ f(\nabla_{W_D}(L)) & \\
W_i^{t+1} &= W_i^t - \lambda\ \nabla_{W_i}(L)^{*} &\hspace{-3cm} i=D-1,\ldots,1
\end{align}

where first the gradient $\nabla_{W_D}(L)$ is transformed using the transformation
function `$f$' and then this transformed gradient is propagated through the rest of the
backward graph. All other gradient vectors $\nabla_{W_i}(L)^{*}$ are computed as is, but
because of the early injection of the transformed gradient $\nabla_{W_D}(L)$, all other
gradient vectors that are connected to the transformed gradient through the chain rule
($\nabla_{W_i}(L)^{*}$,\ \ i=D-1,\ldots,1), gets subsequently altered.

Type I modification is computationally more expensive than Type II modification as it
requires altering the gradients of each and every parameter individually. Type II
modification recomputes gradients at each and every location through the natural flow of
backpropagation. We propose PowerGrad Transform (PGT) which is a type II gradient
modification method. With virtually no additional compute or memory requirement,
PowerGrad Transform enhances classification performance of a convolutional neural
network, by leading the network to arrive at a higher quality optima at convergence, at
which both training and test accuracies improve.

PowerGrad Transform modifies the gradients at the softmax layer, which is the earliest
part of most convolutional neural network's backward computation graph. By changing the
gradient at this stage, all other gradients for all other parameters in the network such
as the linear layer, batchnorm statistics and convolutional layer's filter parameters
are affected. We analyze and explore the behaviour of such alteration both
mathematically and through empirical experiments conducted on networks without
batch-normalization. Experimentally, we see that PGT improved training and test
accuracies for both normalized and non-normalized networks. PGT improves a network's
learning capacity and results in a better fit of the training data.

The following are the major contributions of this paper:

\begin{enumerate}

\item We introduce PowerGrad Transform, which decouples the backward and the forward
passes of neural network training and enables a considerably better fit to the dataset,
as assessed by both training and test accuracy measures. PGT is a performance
enhancement method that alters the gradients in the backward pass before the update step
leading to accelerated training and a significant boost in the network's predictive
performance.

\item We perform theoretical analysis of the properties of the PowerGrad transformation
(section \ref{sec:pgt_prop}) and explore its effect on weight parameters and gradients
(section \ref{sec:pgt_weight_gradients}), logits (section
\ref{sec:pgt_logit_derivative}), (section \ref{sec:Effe}), loss values (section
\ref{sec:Effe}) and class separation (section \ref{sec:pgt_class_separation}).

\item We study experimentally the degenerate behaviour of non-BN networks that is often
seen during the normal training procedure, particularly at larger batch sizes. With PGT,
we see improved gradient behaviour and a decreased likelihood of the weights attaining
degenerate states.

\item We provide complete results from a variety of models (non-BN and BN ResNets,
SE-ResNets) using the ImageNet dataset. It helps the network to improve its learning
capabilities by locating a more optimum convergence point. Additionally, we conduct an
ablation study and compare its impacts to those of regularization approaches such as
label smoothing.

\end{enumerate}

\section{Related Works}
\label{sec:Rela}

We examine techniques that are related to gradient modification and emphasize the
distinction between them and our proposed method.

Gradient Clipping (GC): Gradient Clipping \cite{pascanu2013difficulty}, often used in
natural language processing methods \cite{merity2017regularizing}, is a technique that
involves changing or clipping the gradients with respect to a predefined threshold value
during backward propagation through the network and updating the weights using the
clipped gradients \cite{zhang2019gradient, smith2020generalization}. By rescaling the
gradients, the weight updates are likewise rescaled, significantly reducing the risk of
an overflow or underflow \cite{pascanu2012understanding}. GC can be used for training
networks without batch-normalization.

The formulation is as follows: If $G$ is a gradient vector which is $\partial L /
\partial \theta$ (the gradient of the loss $L$ with respect to the parameters $\theta$).
In clip by value, the gradients are modified as:

\begin{equation}
G \rightarrow
\begin{cases}
\lambda \frac{G}{\norm{G}} & \text{if } \norm{G} > \lambda, \\
G & \text{otherwise}
\end{cases}
\end{equation}

At larger batch sizes, the clipping threshold in GC becomes highly sensitive and
requires extensive finetuning for various models, batch sizes, and learning rates. As we
demonstrate later in our studies, GC is not as effective in improving the performance of
non-normalized networks. AGC performs better than GC in non-normalized networks.
However, we show that PGT outperforms both in such networks.

Adaptive Gradient Clipping (AGC): Adaptive Gradient Clipping \cite{brock2021high} is
developed to further enhance backward pass gradients than what is performed by GC. It
takes into account the fact that the ratio of the gradient norm to the weight norm can
provide an indication of the expected change in a single step of optimization. Here the
normalized gradient is clipped by $\lambda$, which means that large weights can have
large gradients and still the normalized gradient can be within $\lambda$. If it is not
within the threshold, then their gradient is normalized by the ratio of norm of the
gradient and norm of the weight in order to avoid gradient explosion. If $W^l$ is the
weight matrix of the $l^\textit{th}$ layer, and $G^l$ is its gradient in the backward
pass, then the following equation is used to modify the gradient of the $i^\textit{th}$
filter before its update:

\begin{equation}
G_{i}^{l}\rightarrow
\begin{cases}
\lambda \frac{\norm{W_{i}^{l} }_{F}^{*}}{\norm{G_{i}^{l} }_{F}} G_{i}^{l} & if\ \frac{\norm{G_{i}^{l} }_{F}}{\norm{W_{i}^{l} }_{F}^{*}} \  >\ \lambda \\
G_{i}^{l} & otherwise
\end{cases}
\end{equation}

The hyperparameter $\lambda$ of AGC is less sensitive to changes in batch size and depth
as observed in \cite{brock2021high}. However, when applied to vanilla residual networks
(ResNet-18), the performance is still afar from normalized variants as demonstrated by
our experiments. In most cases, AGC is used for training networks without
batch-normalization. Our experiments demonstrate that PGT outperforms AGC, when used
independently. When used together, it can further improve a network's performance
(section \ref{sec:Empi}).

Label Smoothing: Label smoothing, introduced by Szegedy et al.
\cite{szegedy2016rethinking}, utilizes smoothing of the ground truth labels as a method
to impose regularization on the logits and the weights of the neural network.

The formulation is as follows. If $q_i$ is the value at the $i^{th}$ index of the one
hot encoded vector of the ground truth label, then the transformed distribution of the
labels $q_i'$ is: \begin{equation} q_i' = \begin{cases} 1-\epsilon & \text{if } i = y,
\\ \frac{\epsilon}{K-1} & otherwise \end{cases} \end{equation} where $\epsilon$ is a
hyperparameter with a value in $[0,1]$ and $y$ denotes the index of the correct class.

Müller et al. investigates how label smoothing work in \cite{DBLP:conf/nips/MullerKH19}.
Using visualizations of penultimate layer activations, they demonstrate that label
smoothing calibrates the networks predictions and aligns them with the accuracies of the
predictions. Parallel works in label smoothing include Pairwise Confusion
\cite{dubey2018pairwise}, combatting Label Noise \cite{reed2014training}, achieving
regularization through intentional label modification \cite{xie2016disturblabel}.

PowerGrad Transform works in a much different way. PGT does not smooth the ground truth
labels, rather it modifies the gradient by smoothing the predicted probability vector.
As we show in section \ref{sec:Math}, our proposed transformation has the opposite
effect of label smoothing. From our studies, we show that PGT in fact leads to larger
generalization gap between the training and test dataset, although it does lead to an
increase in both training and test accuracies at convergence. Coupled with PGT, label
smoothing can narrow the generalization gap, while retaining the benefits of PGT.

Knowledge Distillation: The formulation of PowerGrad Transform is akin to temperature
based logit suppression of \cite{hinton2015distilling}. Knowledge distillation
\cite{DBLP:conf/nips/BaC14} is a process in which two networks are trained. First a
teacher network is trained on a given dataset and then the soft-labels (the predicted
probabilities) of the teacher network are used to train another network, called the
student network. In similar style of label smoothing, knowledge distillation aims to
generate smooth gradient behaviour by forcing the network not to overpredict on any
single image. As in case of label smoothing, the student network's weights are
automatically penalized if the network assigns a probability value higher than the soft
labels generated by the teacher network. Variants of knowledge distillation include
self-distillation \cite{zhang2019your, wang2021memory}, identical student network
distillation \cite{furlanello2018born}, channel distillation \cite{ge2019distilling},
regularizing wrong predictions \cite{yun2020regularizing} and representation or
embedding based knowledge distillation \cite{aguilar2020knowledge, yao2018deep,
passalis2018unsupervised}. Distillation is applied to various tasks such as model
compression \cite{wang2020real}, face recognition \cite{ge2018low}, enhancing ensemble
performance \cite{zhang2018deep}, interpreting neural networks \cite{liu2018improving}.
Efficacy of distillation is extensivly studied in \cite{cho2019efficacy,
yuan2020revisiting}. Applications of logit suppression include: deep metric learning
\cite{zhai2018classification}, non-parametric neighborhood component analysis
\cite{wu2018improving}, sequence-to-sequence modelling \cite{chorowski2016towards},
reinforcement learning \cite{he2018determining} and face recognition
\cite{liu2017sphereface, wang2018cosface}.

The key difference between PGT and distillation methods is that in the latter, the
transformation is applied in the forward pass, while PGT is a backward pass modification
only. Also, in distillation settings the temperature parameter is a part of the
network's computation graph. In the case of PowerGrad Transform, we directly tamper the
gradients without introducing any change in the forward pass. It is a more generalized
way of arriving at altered gradients in the backward pass. Moreover we show that
modifying the gradients in the backward pass can improve performance in many cases. The
rise in training accuracy indicates that PGT enables the network to learn more from the
training data.

Self-knowledge distillation is used in many papers in different ways, which mainly
focuses on improving the performance of the model by either creating multiple ensemble
student models \cite{zhang2018deep}, using different distorted versions of same data
with two different versions of a single network which guides one another
\cite{xu2019data} or injecting multiple sub-modules in the network to improve its
capacity \cite{zhang2019your}. However none of these methods use normal traditional
training without introducing sub-modules in the architecture. PGT differs from
self-knowledge distillation as it neither introduces any additional sub-modules nor
creates different ensembles to improve the performance of the model. PGT follows the
standard neural network training mechanism with a single change which is altering the
last layer's gradients in the backward pass.

Other techniques of weight update include slot machines \cite{aladago2021slot}, where
random weights are generated and carefully selected during backpropagation in a way that
minimizes the loss function.

\section{PowerGrad Transform}
\label{sec:Powe}

\begin{figure}[t]
\centering
\begin{subfigure}{.5\textwidth}
\centering
\includegraphics[width=0.98\textwidth]{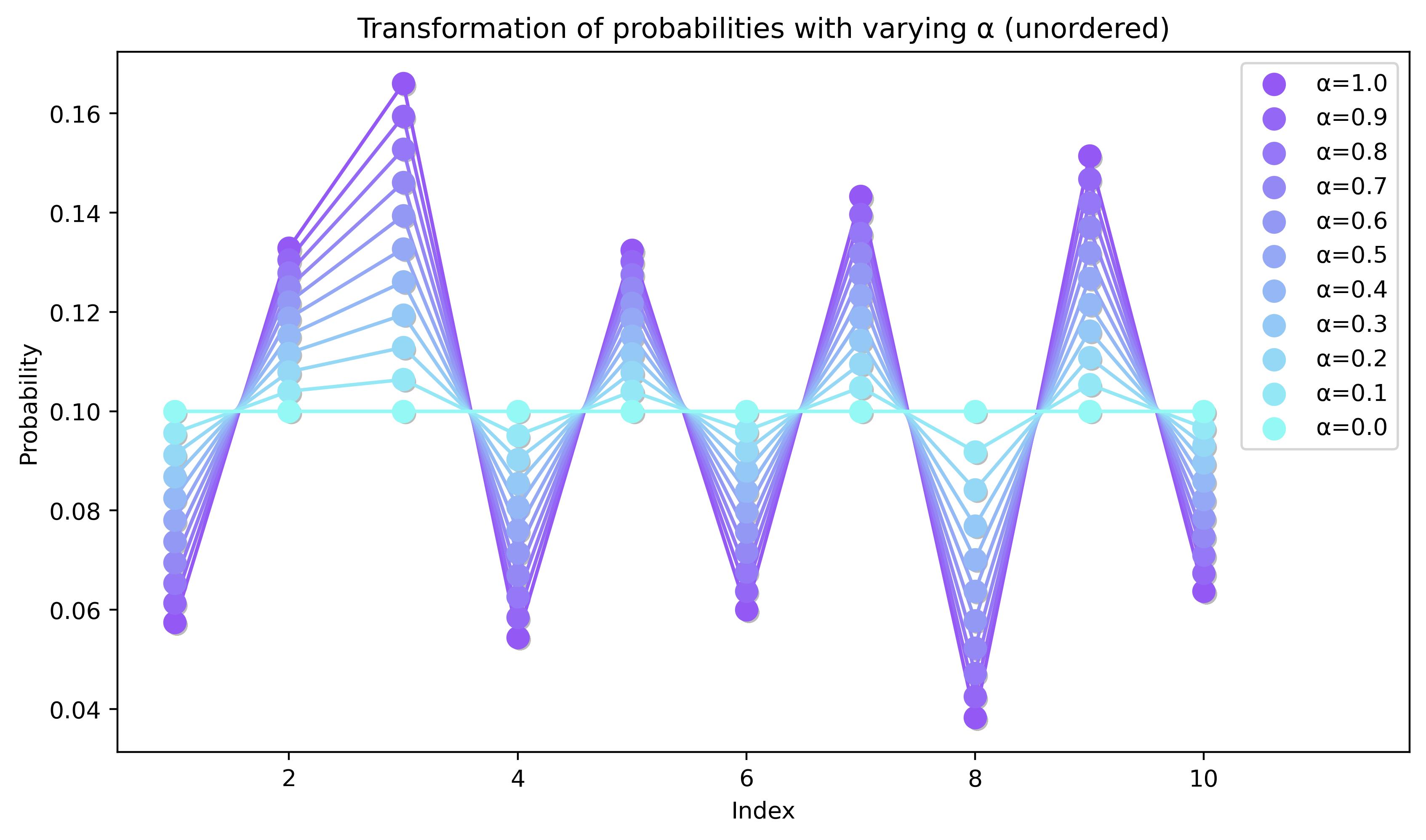}
\caption{Unordered distribution}
\end{subfigure}%
\begin{subfigure}{.5\textwidth}
\centering
\includegraphics[width=0.98\textwidth]{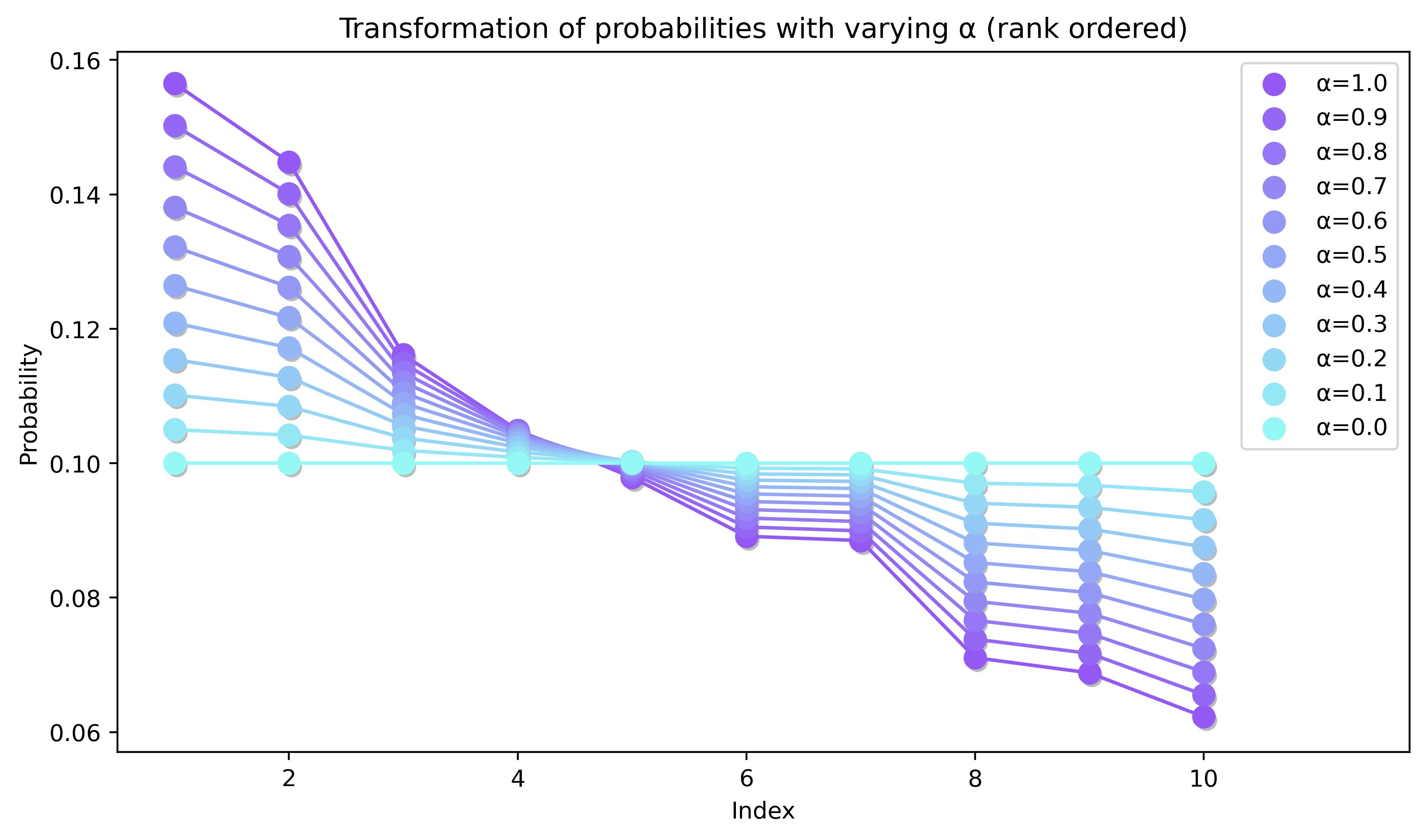}
\caption{Rank ordered distribution}
\end{subfigure}
\caption{ Illustration of the impact of PowerGrad Transform on probability values of an
arbitrary distribution (number of classes, $C=10$). $\alpha=1$ indicates the unmodified
distribution. As $\alpha$ reduces from $1$ to $0$, the probability distributions (left:
unordered or right: ordered) becomes flatter and flatter, approaching the uniform
distribution of $1/C$ for every index. }
\label{fig:prob_plots}
\end{figure}

The PowerGrad Transform technique is described in this section. It is a general
technique and can be applied to any neural network. A neural network with parameters $W$
generates $C$ logits denoted by $z$ for every input vector $x$. $z$ is given as $z=Wx$.
Then a set of probability values $p_i$ are generated from the logits using a softmax
function which is defined as $p_i=\frac{e^{z_i}}{\sum _{j=1}^C e^{z_j}}$. $p_i$ and
$z_i$ represent the predicted probability values and the logits for the $i^{th}$ class
respectively. Following this step, the loss function is invoked and the loss between the
predicted probability values the ground truth labels (which is also a probability
distribution) is calculated. If the loss function is cross-entropy loss, then the value
of the loss is given as $L=-\sum _{i=1}^C q_i \log \left(p_i\right)$ where $q_i$ is the
ground truth label of the $i^{th}$ class for a particular training example. By standard
gradient update rule, we can calculate the gradient of the loss with respect to the
logits which takes the form $\frac{\partial L}{\partial z_i}=p_i-q_i$.

The PowerGrad Transform technique is now described. We introduce a hyperparameter
$\alpha$, which takes a value between $[0, 1]$ and regulates the degree of gradient
modification. Fig. \ref{fig:prob_plots} shows the effects of the transform under
different values of $\alpha$ for different distributions. The PowerGrad Transform method
modifies the predicted probability values in the backward pass as follows:

\begin{equation} \hspace*{3.5cm} p_i'=\frac{p_i^{\alpha }}{\sum _{j=1}^C
p_j^\alpha} \ \ \ \ \ \ \ \ i=1,\dots,C\ \ \ \ \ 0 \leq \alpha\leq 1
\label{transformed_probabilities} \end{equation}

The above transformation changes the gradient of the loss with respect to the logits as
follows:

\begin{equation} \widehat{\frac{\partial L}{\partial z_i}}=p_i'-q_i
\label{PGT_logit_derivative}
\end{equation}

The rest of the backward pass proceeds as usual. We denote the original probability
distribution as $P$ (with values $p_i$ at the $i^{th}$ index) and the transformed
distribution as $P'$ (with values $p_i'$ at the $i^{th}$ index).

A code snippet of the algorithm implemented using PyTorch \cite{NEURIPS2019_9015} is
given in the appendix. In PyTorch, the available method to modify gradients in the
backward pass is using the \codeword{register\_hook} function which is called in the
\codeword{forward} function. We explore the effect of this change from a theoretical
standpoint in section \ref{sec:Math}. In section \ref{sec:Expe}, we offer empirical data
and experiments to support the proposed method.

\section{Analysis of the PowerGrad transformation}
\label{sec:Math}

In this section, we first describe the properties of the  PowerGrad Transform (PGT) and
then we highlight how these suitable properties of PGT helps the neural network to
improve the performance. We use the same setup as described in section \ref{sec:Powe}.
To explore the properties of PGT, we start by investigating the effect of the transform
on the softmax probabilities.

\subsection{Properties of PGT}
\label{sec:pgt_prop}

\textbf{Lemma 1.} For any arbitrary probability distribution $P$ with probability values
given by $p_i$ for $i=1,\dots,C$, the corresponding transformed probability values
$p_i'$ given by [Eq. \ref{transformed_probabilities}] has a threshold $\Big(\sum
_{j=1}^C p_j^{\alpha}\Big)^{\frac{1}{\alpha-1}}$ and

\begin{equation} \begin{split} p_i' \geq p_i & \text{,\ \ if } p_i \leq
\Big(\sum _{j=1}^C p_j^{\alpha }\Big)^{\frac{1}{\alpha-1}} \\ \hspace*{1cm}
p_i' < p_i & \text{,\ \ if } p_i > \Big(\sum _{j=1}^C
p_j^{\alpha}\Big)^{\frac{1}{\alpha-1}} \end{split} \label{eqn:threshold}
\end{equation}

\textit{Proof.} The proof follows directly from the definition of $p_i'$. We notice that
if $p_i' < p_i$, we get:

\begin{proof}[\unskip\nopunct] \begin{alignat}{2} &\ \ \frac{p_i^{\alpha }}{\sum
_{j=1}^C p_j^{\alpha}} < p_i \\ \Rightarrow &\ p_i^{\alpha -1}<\sum _{j=1}^C p_j^{\alpha
} \\ \Rightarrow &\ p_i > \Big(\sum _{j=1}^C p_j^{\alpha}\Big)^{\frac{1}{\alpha-1}}
\end{alignat} \end{proof}

We call this threshold, the \textit{stationary threshold}. The stationary threshold is
that value of $p_i$ that does not change after the transformation. Therefore, when $p_i$
is greater than the \textit{stationary threshold}, $p_i' < p_i$.

\textbf{Proposition 1.} At $\alpha=0$, the stationary threshold equals $1/C$ and all
values of the transformed distribution $p_i'$ reduces reduces to the uniform
distribution for $i=1,\dots,C$,.

\textit{Proof.} From Eq. (\ref{eqn:threshold}), we see that the stationary threshold at
$\alpha=0$ is $1/C$. Also, following from the definition of the transformed
probabilities (Eq. \ref{transformed_probabilities}) we conclude that if $\alpha=0$, then
all values of $p_i'$ are $1/C$. Therefore the transformed distribution at $\alpha=0$ is
a uniform distribution.

Since we have established that values of $p_i$ which are greater than the stationary
threshold decreases and move down towards the stationary threshold, and values in $p_i$
lower than the stationary threshold moves up towards the stationary threshold, this
transformation makes the distribution more uniform (i.e. it smooths out the actual
distribution) as $\alpha$ is decreased from $1$ and down towards $0$. This final
observation we prove in the following theorem.

\textbf{Theorem 1.} For any arbitrary probability distribution $P$ with probability
values $p_i$ for $i=1,\dots,C$, the stationary threshold of the transformed distribution
$P'$ with probability values $p_i'=\frac{p_i^{\alpha}}{\sum _{j=1}^C p_j^\alpha}, 0 \leq
\alpha \leq 1$ is a monotonically non-decreasing function with respect to $\alpha$.

\textit{Proof.} To prove monotonicity, we first compute the gradient of the stationary
threshold with respect to the variable in concern, $\alpha$.

\begin{alignat}{2} & \frac{\partial }{\partial \alpha} \left(\sum _{j=1}^c
p_j^{\alpha}\right){}^{\frac{1}{\alpha -1}}\ \ = \left(\sum _{j=1}^c
p_j^{\alpha}\right){}^{\frac{1}{\alpha -1}} \left(\frac{\sum _{j=1}^c p_j^{\alpha } \log
\left(p_j\right)}{(\alpha -1) \sum _{j=1}^c p_j^{\alpha}}-\frac{\log \left(\sum _{j=1}^c
p_j^{\alpha}\right)}{(\alpha -1)^2}\right) \\
& = \frac{1}{\alpha  (\alpha -1)^2} \left(\sum _{j=1}^c p_j^{\alpha
}\right){}^{\frac{1}{\alpha -1}} \left(\frac{(\alpha -1) \sum _{j=1}^C p_j^{\alpha }
\log \left(p_j^{\alpha}\right)}{\sum _{j=1}^c p_j^{\alpha }}-\alpha \log \left(\sum
_{j=1}^c p_j^{\alpha }\right)\right) \label{derivative} \end{alignat}

If $a_1,\dots ,a_n$ and $b_1,\dots ,b_n$ are non-negative numbers, then using the log
sum inequality, \\ we get $\sum _{j=1}^n a_j \log \left(\frac{a_j}{b_j}\right)\geq
\left(\sum _{j=1}^n a_j\right) \log \left(\frac{\sum _{j=1}^n a_j}{\sum _{j=1}^n
b_j}\right)$. Setting $a_j=p_j^\alpha$ and $b_j=1$, we get the following lower bound
\begin{equation} \sum _{j=1}^C p_j^{\alpha } \log
\left(p_j^{\alpha }\right)\geq \left(\sum _{j=1}^C
p_j^{\alpha}\right) \log \left(\frac{1}{C} \sum _{j=1}^C
p_j^{\alpha}\right) \label{lower_bound1} \end{equation}

Substituting (\ref{lower_bound1}) in (\ref{derivative}), we get:

\begin{equation} \frac{\partial }{\partial \alpha} \left(\sum
_{j=1}^c p_j^{\alpha}\right){}^{\frac{1}{\alpha -1}}\ \ \geq
\frac{1}{\alpha  (\alpha -1)^2} \left(\sum _{j=1}^c
p_j^{\alpha}\right){}^{\frac{1}{\alpha -1}} \left((1-\alpha )
\log (C)-\log \left(\sum _{j=1}^C p_j^{\alpha }\right)\right)
\label{lb_subs} \end{equation}

$p^{\alpha}$ is concave, and so by Jensen's inequality we get the following upper bound
for the second term:

\begin{equation} \left(\frac{1}{C}{\sum _{j=1}^C
p_j}{}\right){}^{\alpha }\ \ \geq\ \ \frac{1}{C}{\sum _{j=1}^C
p_j^{\alpha }}{} \end{equation}

\begin{equation}\Rightarrow\ \ \log \left(\sum _{j=1}^C p_j^{\alpha}\right)\leq
(1-\alpha ) \log (C) \label{upper_bound} \end{equation}

Substituting (\ref{upper_bound}) in (\ref{lb_subs}),

\begin{proof}[\unskip\nopunct] \begin{equation} \frac{\partial}{\partial \alpha
}\left(\sum _{j=1}^c p_j^{\alpha}\right){}^{\frac{1}{\alpha -1}}\ \ \geq \ \ 0
\end{equation} \end{proof}

We conclude from the analysis that the stationary threshold is a monotonic
non-decreasing function with respect to $\alpha$. Also the derivative of PGT function
with respect to the true probabilities is non-negative which in turn means that the
transformation is an order-preserving map. All values greater than the threshold move
towards the threshold after transformation and all values below the threshold also move
towards the threshold, and the threshold itself moves monotonically towards $1/C$ as
$\alpha$ is decreased from $1$ to $0$. This concludes that the transformation smooths
the original distribution.

\subsection{PGT restricts the partial derivative of the loss w.r.t. each logit from becoming too small or too high}
\label{sec:pgt_logit_derivative}

We know that neural networks uses the softmax function to generate prediction
probabilities from logits in multi-class classification tasks. If we use cross-entropy
loss to train the network then the partial derivative of the loss w.r.t. $i^{th}$ logit
is dependent on the value of the predicted probability and the value of the class label
(either 0 or 1) of the $i^{th}$ logit. 

\begin{equation}
\label{actual_logit_derivative}
\frac{\partial L}{\partial z_i} = p_i -q_i
\end{equation}
In general, the range of the $|\frac{\partial L}{\partial z_i}|$ is [$\delta$,
$\epsilon$] where, $\delta\rightarrow 0$ and $\epsilon\rightarrow 1$ in traditional
training (Eq. \ref{actual_logit_derivative} procedures of neural networks. Here, we
propose a new method of neural network training which alters the gradients at the time
of backward pass by modifying the predicted probability vector using PGT (Eq.
\ref{PGT_logit_derivative}) such that the directional derivative for each logit does not
become too small or too large,

With inclusion of PGT in the last layer, the range of the $\widehat{|\frac{\partial
L}{\partial z_i}|}$ becomes [$\delta^+$, $\epsilon^-$] where, $\delta^+>\delta$ and
$\epsilon^-<\epsilon$

Now, we demonstrate how PGT restricts the magnitude of the directional derivative for
each logit. For a training example, a neural network can assign the highest probability
to either (i) a wrong class or to the (ii) correct class. 

\textbf{(i) Wrong Class Predicted:}
If the network assigns the highest probability for any class except the the true class,
then wrong class assignment happens. We analyze the effect on gradient due to PGT for
this case. There are three possibilities

(a) Let, $i^{th}$ class be the class for which the network assigns low probability, but
it is the true class, then

$p_i < p_{th} \Rightarrow p'_i > p_i \Rightarrow \widehat{\frac{\partial L}{\partial
z_i}} = p'_i -1 >  p_i -1 $ ; where $p_i\rightarrow 0$ \& $q_i = 1$

So; $|\widehat{\frac{\partial L}{\partial z_i}}| = |p'_i -1| <  |p_i -1| \Rightarrow
|\widehat{\frac{\partial L}{\partial z_i}}| <\epsilon $ where; $\epsilon\rightarrow 1$

(b) Let, $j^{th}$ class be the class (not true class) for which the network assigns the
highest probability, then

$p_j > p_{th} \Rightarrow p'_j < p_j \Rightarrow \widehat{\frac{\partial L}{\partial
z_j}} = p'_j <  p_j$ ; where $p_j\rightarrow 1$ \& $q_j = 0$

So, $|\widehat{\frac{\partial L}{\partial z_j}}| <\epsilon$ where; $\epsilon\rightarrow
1$

(c) Let $k$ denote the index for all those classes for which the predicted probability
is low and also the ground truth class label is $0$, then

$p_k < p_{th} \Rightarrow p'_k > p_k \Rightarrow \widehat{\frac{\partial L}{\partial
z_k}} = p'_k >  p_k$ ; where $p_k\rightarrow 0$ \& $q_k = 0$

So, $|\widehat{\frac{\partial L}{\partial z_k}}| >\delta$ where; $\delta\rightarrow$ 0

\textbf{(ii) Correct Class Predicted:}
For a given training example, if the network assigns the highest probability to the true
class then it assigns correct class to the training example. Now we observe the effect
on gradient due to PGT for this case. There are two possibilities,

(a) Let, $i^{th}$ class be the class for which the network assigns the highest
probability and it is also the true class, then

$p_i > p_{th} \Rightarrow p'_i < p_i \Rightarrow \widehat{\frac{\partial L}{\partial
z_i}} = p'_i -1 <  p_i -1 $ ; where $p_i\rightarrow 1$ \& $q_j = 1$

So; $|\widehat{\frac{\partial L}{\partial z_i}}| = |p'_i -1| >  |p_i -1|  \Rightarrow
|\widehat{\frac{\partial L}{\partial z_i}}| >\delta $ where; $\delta\rightarrow 0$

(b) Let, $k^{th}$ index is used for all those classes for which the predicted
probability is low and also the actual label value $0$, then

$p_k < p_{th} \Rightarrow p'_k > p_k \Rightarrow \widehat{\frac{\partial L}{\partial
z_k}} = p'_k >  p_k$ ; where $p_k\rightarrow 0$ \& $q_k = 0$

So, $|\widehat{\frac{\partial L}{\partial z_k}}| >\delta$ where; $\delta\rightarrow 0$

After analyzing all these cases ((i)a,b,c \& (ii)a,b), we are able to deduce that PGT
helps to restrict the directional derivative to be within limit such that for every
logit/direction the directional derivative is not very small or very large. 

\subsection{Effect of PGT on weight gradients}
\label{sec:pgt_weight_gradients}

\begin{figure}[t]
\centering
\includegraphics[scale=1.0,width=0.5\columnwidth]{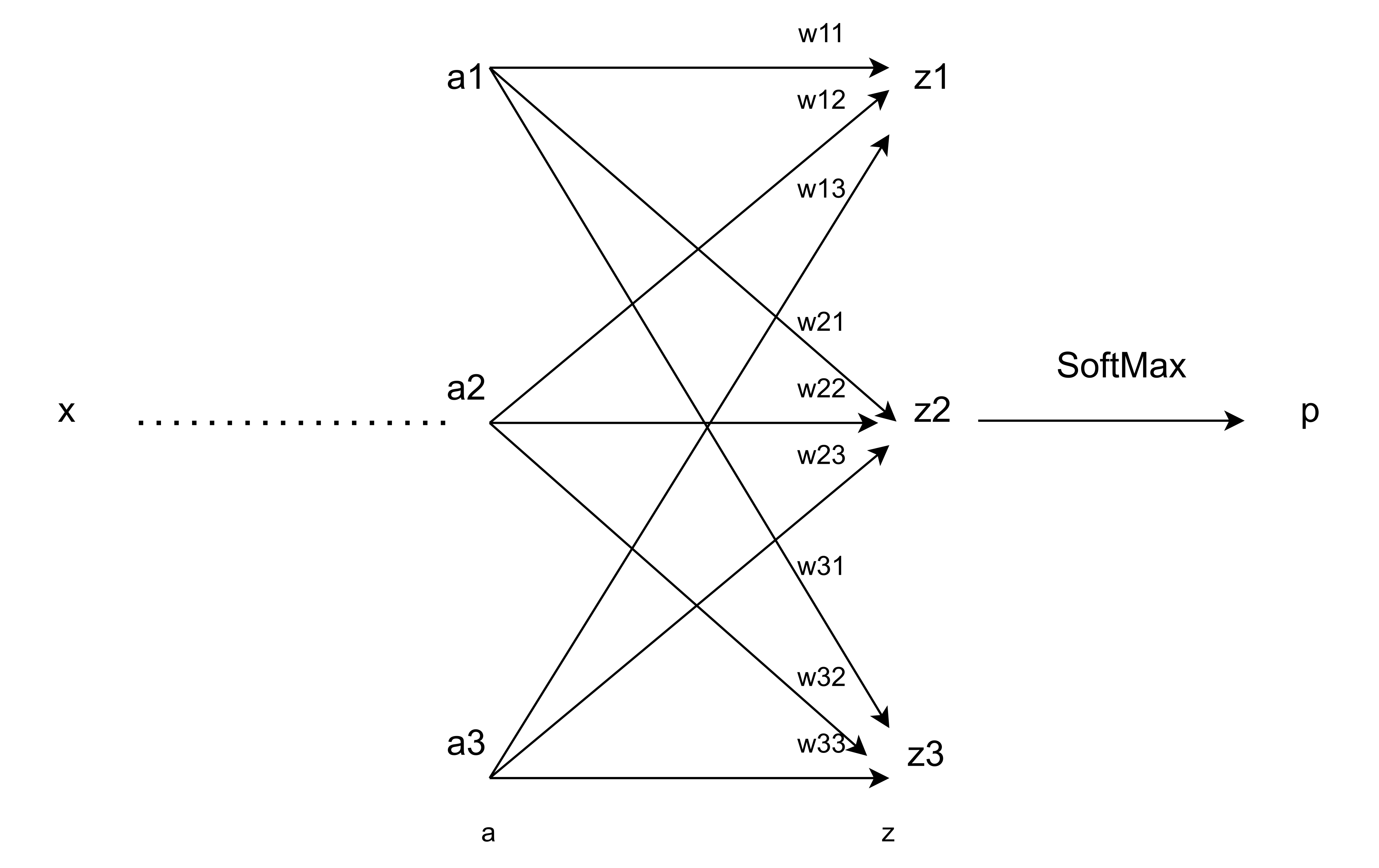}
\caption{Fully connected layer of a convolutional neural network} 
\label{fig:simplenn}
\end{figure}

In this section, we discuss the impact of PGT on weight gradients. Let the input and
output of the last convolutional layer of a convolutional neural network be \textbf{x}
and \textbf{a} respectively (Fig. \ref{fig:simplenn}). The last layer which is a fully
connected layer, produces logits \textbf{z} by combining \textbf{a} with weights and
thereafter the network generates the predicted probability vector \textbf{p} from the
logits using softmax activation function. We note that a$_i \geq 0$ $ \forall i$  as,
\textbf{a} is the output of ReLU activation functions.

Here, $z_i = \sum_{j=1}^{C}w_{ij}a_j$, so $\frac{\partial L}{\partial w_{ij}} =
a_j\frac{\partial L}{\partial z_{i}} $. However, if we apply PGT while training the
neural network, then $\widehat{\frac{\partial L}{\partial w_{ij}}}
=a_j\widehat{\frac{\partial L}{\partial z_i}}$

If the activation of the $j^{th}$ neuron is zero, i.e. $a_j = 0$ then there is zero
gradient for all weights which acts on $a_j$ both for normal training and training with
PGT, as $a_j$ has no contribution in the computation of logits. However, the interesting
thing is to observe the effect of PGT when $a_j > 0$. As, the activation values $(a_j)$
are positive, so $ \widehat{\frac{\partial L}{\partial z_i}}  > \frac{\partial
L}{\partial z_{i}} \Rightarrow$ $\widehat{\frac{\partial L}{\partial w_{ij}}} >
\frac{\partial L}{\partial w_{ij}} \forall j \in (1,2,\ldots C)$ and similar
relationship for $a_j < 0$. With PGT, the weight updating rule for $(t+1)^{th}$
iteration then becomes,
\begin{equation}
\label{weight_update_PGT}
w_{ij}^{t+1} = w_{ij}^{t} - \eta \widehat{\frac{\partial L}{\partial w_{ij}^{t}}}
\end{equation}
The weight update equation using PGT, can also be viewed in terms of the following
equation,
\begin{equation}
\label{weight_update_final_PGT}
w_{ij}^{t+1} = w_{ij}^{t} - \frac{\eta}{C_{PGT}} \frac{\partial L}{\partial w_{ij}^{t}}
\end{equation}

For neural networks, the exponential nature of the softmax function produces sharp
distribution for predicted probability vector. For example, if \textbf{z} is $[20, 30,
10]$ then \textbf{p} is equal to $[4\times 10^{-5},  0.99995, 2\times10^{-9}]$. When the
network is at the initial phase of training, most of the training examples are
misclassified. Under this scenario, PGT helps to update the weights in a better manner
than traditional gradients. Suppose the label vector \textbf{q} is $[1, 0, 0]$ then it
indicates that $z_1$ is the logit corresponding to ground truth class and $z_2$ is the
logit where the network has assigned the highest probability. Similarly, $z_3$ is the
logit where the network has assigned a low probability as well as the label vector is
also $0$ for this logit. Here $\frac{\partial L}{\partial w_{1j}^{t}} = \epsilon a_j > 0
\Rightarrow$ $ \widehat{\frac{\partial L}{\partial w_{1j}^{t}}} < \frac{\partial
L}{\partial w_{1j}^{t}} \Rightarrow C_{PGT} > 1$ $\forall j \in (1, 2,\ldots C)$.
Advanced gradient updation algorithms like AdaGrad \cite{duchi2011adaptive} reduces the
gradient in the direction where the directional derivative for a weight is high. It
reduces the learning rate for the weight so that the gradient in that direction is not
too large in order to make the update process less volatile. The formulation of PGT also
implicitly reduces the partial derivative in the same manner for the high derivative
directions, the only difference with the AdaGrad is that $C_{PGT}$ depends only on the
current iteration's gradient and not on previous iterations like AdaGrad. Similarly we
can observe that for this example, $\frac{\partial L}{\partial w_{3j}^{t}} = \delta a_j
\approx 0 \Rightarrow$ $  \widehat{\frac{\partial L}{\partial w_{3j}^{t}}} >
\frac{\partial L}{\partial w_{3j}^{t}} \Rightarrow C_{PGT} < 1$ $\forall j \in (1,
2,\ldots C)$. As directional derivative for the weights associated with third logit is
extremely low, PGT helps to increase the derivative in these directions so that the
training process does not become too slow.

\subsection{Effect of PGT on class separation}
\label{sec:pgt_class_separation}

During final iterations of training, we analyze the iteration-wise gradient behaviour
where the network assigns the highest probability to the correct class for a given
training example; we deduce the effects of PGT in the next immediate training iteration.
For example, if \textbf{z} is $[30, 20, 10]$ then \textbf{p} is $[0.99995, 4\times
10^{-5}, 2\times10^{-9}]$ and \textbf{q} is $[1, 0, 0]$. There is no change in the
weight update due to PGT when $a_j = 0$. However, when $i \neq 1$ and $a_j > 0$,
$|\widehat{\frac{\partial L}{\partial w_{1j}^{t}}}| > |\frac{\partial L}{\partial
w_{1j}^{t}}|$ and $\widehat{\frac{\partial L}{\partial w_{ij}^{t}}} > \frac{\partial
L}{\partial w_{ij}^{t}}$. Here in this example $\frac{\partial L}{\partial z_{1}^{t}} <
0 \Rightarrow \frac{\partial L}{\partial w_{1j}^{t}} < 0 \Rightarrow$
$\widehat{w_{1j}^{t+1}} > w_{1j}^{t+1} > w_{1j}^{t} \forall j$. We also know that
$z_i^{t+1} = \sum_{j=1}^{C}w_{ij}^{t+1}a_j$ and $\widehat{z_i^{t+1}} =
\sum_{j=1}^{C}\widehat{w_{ij}^{t+1}}a_j$. It indicates that $\widehat{z_{1}^{t+1}} >
z_{1}^{t+1} > z_{1}^{t}$ as all the weights associated with first logit is larger while
using PGT in $(t+1)^{th}$ iteration. Similarly, for the non-true classes
$\widehat{w_{ij}^{t+1}} < w_{ij}^{t+1} < w_{ij}^{t}$ when $i \neq 1$ $\forall j$ as
$\frac{\partial L}{\partial z_{i}^{t}} > 0$. So in $(t+1)^{th}$ iteration,
$\widehat{z_{i}^{t+1}} < z_{i}^{t+1} < z_{i}^{t}$ when $i \neq 1$. We observe that
$dist(\widehat{z_{1}^{t+1}}, \widehat{z_{i}^{t+1}}) > $  $dist(z_{1}^{t+1},
z_{i}^{t+1})$ where the first logit corresponds to the true class and all other logits
indexed by $i$ are non-true classes. Therefore the distance between correct and
incorrect class logits increases due to PGT and this leads to better class separation.

\section{Experiments}
\label{sec:Expe}

\subsection{Experiments on convolutional neural networks (with batch-normalization)}
\label{sec:bn}

\begin{figure}[!t]
\centering
\begin{subfigure}{.33\textwidth}
\centering
\includegraphics[width=0.98\textwidth]{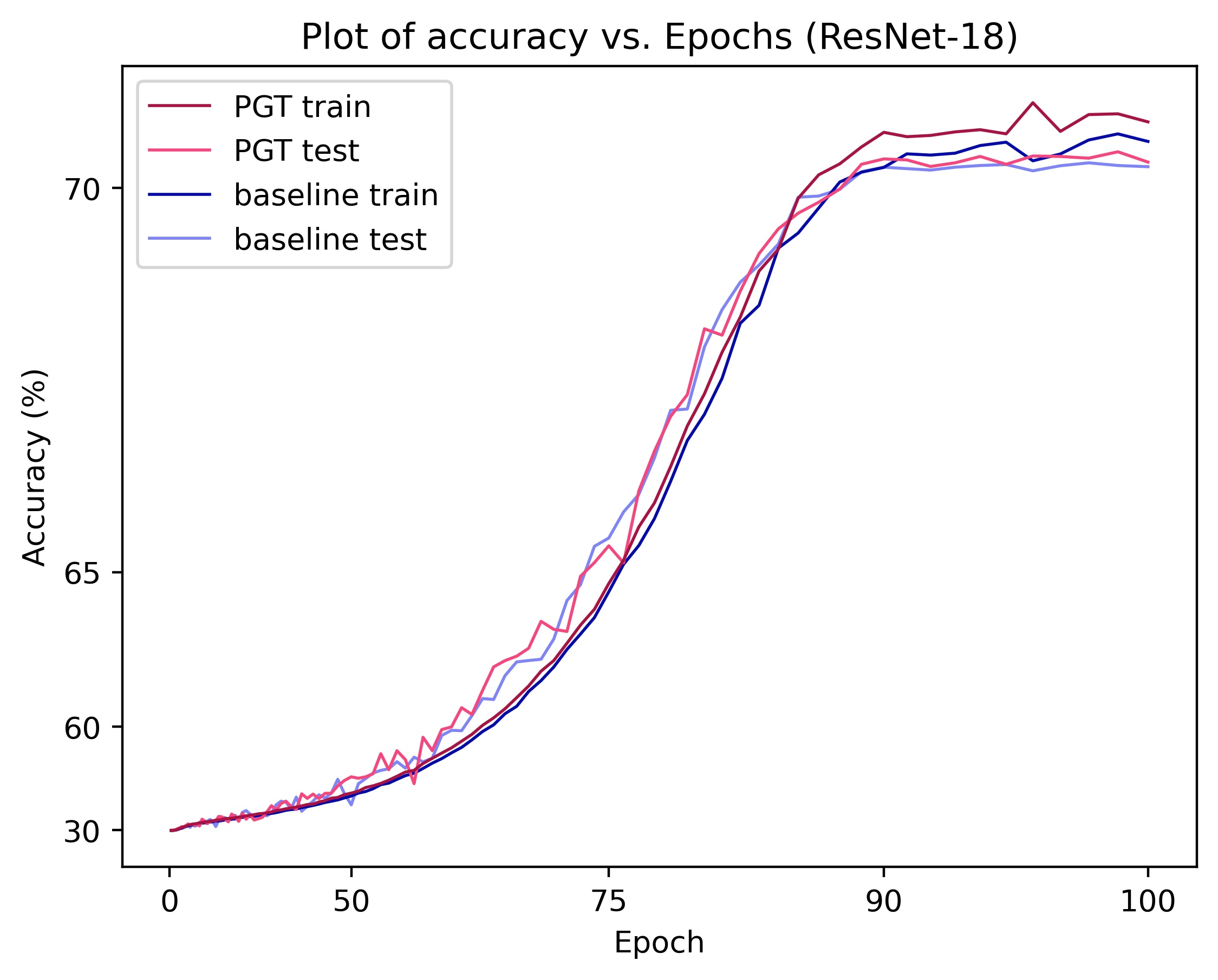}
\caption{Plot of training and test accuracies (ResNet-18)}
\end{subfigure}%
\begin{subfigure}{.33\textwidth}
\centering
\includegraphics[width=0.98\textwidth]{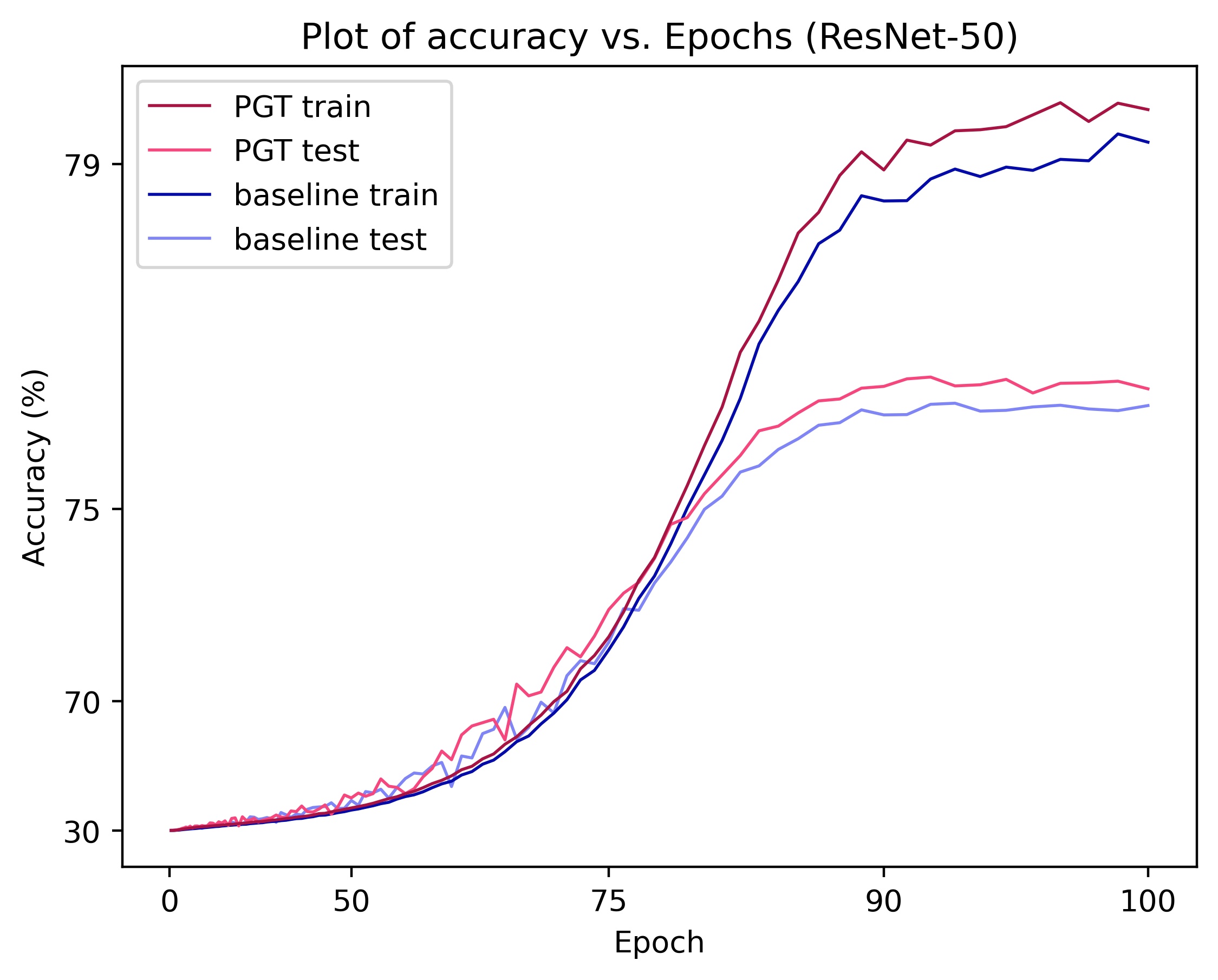}
\caption{Plot of training and test accuracies (ResNet-50)}
\end{subfigure}%
\begin{subfigure}{.33\textwidth}
\centering
\includegraphics[width=0.98\textwidth]{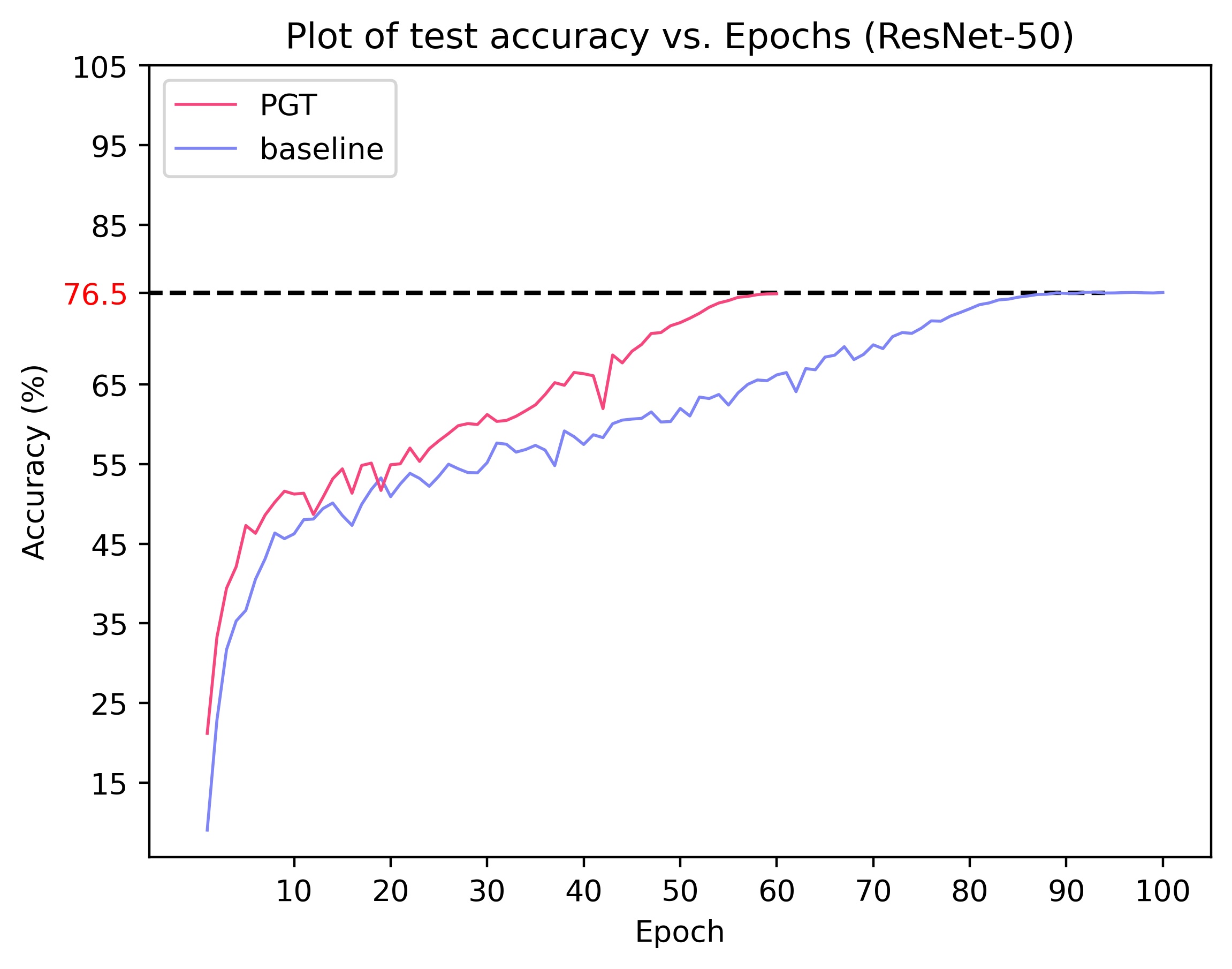}
\caption{Plot of test accuracies (ResNet-50)}
\end{subfigure}
\caption{ Log-linear plots of training and test accuracies and comparison with baseline
of batch-normalized variants: \textbf{(a)} ResNet-18 ($\alpha=0.25$), \textbf{(b)}
ResNet-50 ($\alpha=0.3$). For better visualization of accuracies obtained in the final
iterations, we plot both axes in log scale. PowerGrad Transform improves upon the
baseline in all cases. The corresponding accuracies are mentioned in Table
\ref{tab:imagenet_table}. \textbf{(c)} Training speed comparison between PGT ($60$
epochs) and baseline ($100$ epochs). They both converge to the same test accuracy
($76.5\%$) on ImageNet-1K. PGT's accelerated training saves $40\%$ of the epoch budget.
}
\label{fig:metrics}
\end{figure}

\begin{table}[!t]
\centering
\caption{ Results and comparison table for CNN networks trained on ImageNet-1K (sorted
in descending order w.r.t. the last column denoting test accuracy improvements over
baseline). Best training and test accuracies are highlighted in red and blue
respectively. Accuracy differences are highlighted in yellow. In all experiments, we
observe that PGT positively improves both training and test accuracies. We also observe
that deeper networks such as SE-ResNet-50 and ResNet-50 accrue higher gains over the
baselines as compared to shallow networks like ResNet-18. }
\label{tab:imagenet_table}
\scalebox{0.7}{
\begin{tabular}{cccccccc}
\multirow{2}{*}{\textbf{Model}} & \multirow{2}{*}{\textbf{Scheduler}} &
\multirow{2}{*}{\textbf{Method}} & \textbf{PowerGrad} & \textbf{\hspace{-0.65cm} Train}
& \textbf{Train} & \textbf{\hspace{-0.65cm} Test} & \textbf{Test} \\
& & & \textbf{Transform ($\alpha$)} & \textbf{Accuracy (\%)} & \textbf{Diff (\%)} &
\textbf{Accuracy (\%)} & \textbf{Diff (\%)} \\
\midrule
\multirow{2}{*}{SE-ResNet-50} & \multirow{2}{*}{Cosine} & Baseline & - & 81.5 &
\textcolor{olive}{\multirow{2}{*}{\textbf{+0.97}}} & 77.218 &
\textcolor{olive}{\multirow{2}{*}{\textbf{+0.734}}} \\
& & PGT & 0.3 & \textcolor{red}{\textbf{82.47}} & & \textcolor{blue}{\textbf{77.952}} &
\\
\midrule
\multirow{2}{*}{ResNet-50} & \multirow{2}{*}{Cosine} & Baseline & - & 79.18 &
\textcolor{olive}{\multirow{2}{*}{\textbf{+0.5}}} & 76.56 &
\textcolor{olive}{\multirow{2}{*}{\textbf{+0.656}}} \\
& & PGT & 0.05 & \textcolor{red}{\textbf{79.68}} & & \textcolor{blue}{\textbf{77.216}} &
\\
\midrule
\multirow{2}{*}{ResNet-50} & \multirow{2}{*}{Step} & Baseline & - & 78.99 &
\textcolor{olive}{\multirow{2}{*}{\textbf{+0.57}}} & 75.97 &
\textcolor{olive}{\multirow{2}{*}{\textbf{+0.524}}} \\
& & PGT & 0.05 & \textcolor{red}{\textbf{79.56}} & & \textcolor{blue}{\textbf{76.494}} &
\\
\midrule
\multirow{2}{*}{ResNet-101} & \multirow{2}{*}{Cosine} & Baseline & - & 82.29 &
\textcolor{olive}{\multirow{2}{*}{\textbf{+0.81}}} & 77.896 &
\textcolor{olive}{\multirow{2}{*}{\textbf{+0.362}}} \\
& & PGT & 0.3 & \textcolor{red}{\textbf{83.1}} & & \textcolor{blue}{\textbf{78.258}} &
\\
\midrule
\multirow{2}{*}{SE-ResNet-18} & \multirow{2}{*}{Cosine} & Baseline & - & 71.42 &
\textcolor{olive}{\multirow{2}{*}{\textbf{+0.18}}} & 71.09 &
\textcolor{olive}{\multirow{2}{*}{\textbf{+0.346}}} \\
& & PGT & 0.25 & \textcolor{red}{\textbf{71.6}} & & \textcolor{blue}{\textbf{71.436}} &
\\
\midrule
\multirow{2}{*}{ResNet-18} & \multirow{2}{*}{Step} & Baseline & - & 69.95 &
\textcolor{olive}{\multirow{2}{*}{\textbf{+0.35}}} & 69.704 &
\textcolor{olive}{\multirow{2}{*}{\textbf{+0.14}}} \\
& & PGT & 0.25 & \textcolor{red}{\textbf{70.3}} & & \textcolor{blue}{\textbf{69.844}} &
\\
\midrule
\multirow{2}{*}{ResNet-18} & \multirow{2}{*}{Cosine} & Baseline & - & 70.38 &
\textcolor{olive}{\multirow{2}{*}{\textbf{+0.15}}} & 70.208 &
\textcolor{olive}{\multirow{2}{*}{\textbf{+0.09}}} \\
& & PGT & 0.25 & \textcolor{red}{\textbf{70.53}} & & \textcolor{blue}{\textbf{70.298}} &
\\
\end{tabular}}
\end{table}

We perform experiments on different variants ResNets using the ImageNet-1K dataset
\cite{deng2009imagenet}. All models are trained on four V100 GPUs with a batch size of
$1024$. We utilize a common set of hyperparameters for all experiments, which are as
follows: 100 epoch budget, 5 epochs linear warmup phase beginning with a learning rate
of $4\times 10^{-4}$ and ending with a peak learning rate of $0.4$, a momentum of $0.9$
and weight decay of $5\times 10^{-4}$, the SGD Nesterov optimizer and mixed precision.
In our studies, we employ either a step scheduler (dividing the learning rate by $10$ at
the $30^{th}$, $60^{th}$, and $90^{th}$ epochs) or a cosine decay scheduler
\cite{loshchilov2016sgdr}. We find $\alpha=0.25$ and $\alpha=0.05$ to be good choices
for ResNet-18 and ResNet-50, though larger values such as $\alpha=0.3$ also have good
performance as well. The experimental results are shown in Table
\ref{tab:imagenet_table}, and we mention the value of the PGT hyperparmeter ($\alpha$)
in each experiment. We explore different values of $\alpha$ in section \ref{sec:Abla}
and provide detailed grid plots in the supplementary section.

In our experiments with Squeeze-and-Excitation variant of ResNet-50 i.e.
SE-ResNet-50\cite{hu2018squeeze}, we observe significant improvements; a $0.97\%$ boost
in training accuracy and a $0.734\%$ increase in test accuracy. In our experiments with
ResNet-50, we find an increase of $0.5\%$ and $0.656\%$ performance enhancement over the
cosine scheduler baseline for training and testing respectively. The corresponding
improvement over the step scheduler baseline for ResNet-50 is $0.57\%$ (training) and
$0.524\%$ (testing). ResNet-101 sees a higher improvement in training fit, $0.81\%$ to
be exact, while the improvement over the test set is $0.362\%$. Smaller networks such as
SE-ResNet-18 and ResNet-18 sees accuracy boosts which are smaller but nevertheless
positive. Therefore we conclude that with PGT, the consistent improvements in training
accuracies across all cases is because the networks train better and arrive at better
optimas during convergence. Per epoch training and test accuracy plots of ResNet-18 and
ResNet-50 (both with and without PGT) are shown in Fig. \ref{fig:metrics}(a, b). For
better visualization of the top end of the plots, we scale both axes in log scale.
Because both training and test accuracies improve, it indicates that PowerGrad Transform
allows the network to learn better representations\footnote{Reproducible code, training
recipes for the experiments, pretrained checkpoints and training logs are provided at:
\\
\url{ https://github.com/bishshoy/power-grad-transform }. Code is adapted from
\cite{rw2019timm}.}.

These improvements are obtained with the number of epochs being kept the same as that of
the baseline model. Alternatively, practitioners can apply PGT and have the networks
converge to accuracy levels comparable to the baseline with a significantly reduced
epoch budget. In Fig. \ref{fig:metrics}(c), we illustrate one such instance in which we
train a ResNet-50 to its cosine scheduler baseline in just $60$ epochs with PGT, and
match its baseline performance that is acquired in $100$ epochs. This hints to the
observation that PGT not only makes the network perform better but also speeds up
training. Because of the acceleration in training that is made possible by PGT, the
total amount of training time can be decreased by $40\%$.

\subsection{Empirical studies on networks without Batch Normalization}
\label{sec:Empi}

This section empirically examines some of the issues that occur when networks are
trained without normalization layers and we provide insight into the effects of
PowerGrad transform on such networks. We use ResNet-18 \cite{he2016deep} as the
foundation model for all empirical trials owing to its popularity among deep learning
practitioners and the relative simplicity with which it can be trained on a big dataset
such as ImageNet-1K \cite{russakovsky2015imagenet}. We found deeper networks (such as
ResNet-34 and ResNet-50 models) are impossible to train without BatchNormalization
because of increased depth. So in order to conduct our empirical experiments, we solely
focus on the ResNet-18 architecture. For our experiments, we designate different layers
with their corresponding layer indices. We provide a detailed layer-wise index list of
ResNet-18 in the supplementary section.

\begin{figure}[t]
\centering
\captionsetup{font=scriptsize}

\begin{subfigure}[t]{0.32\textwidth}
\includegraphics[width=\textwidth]{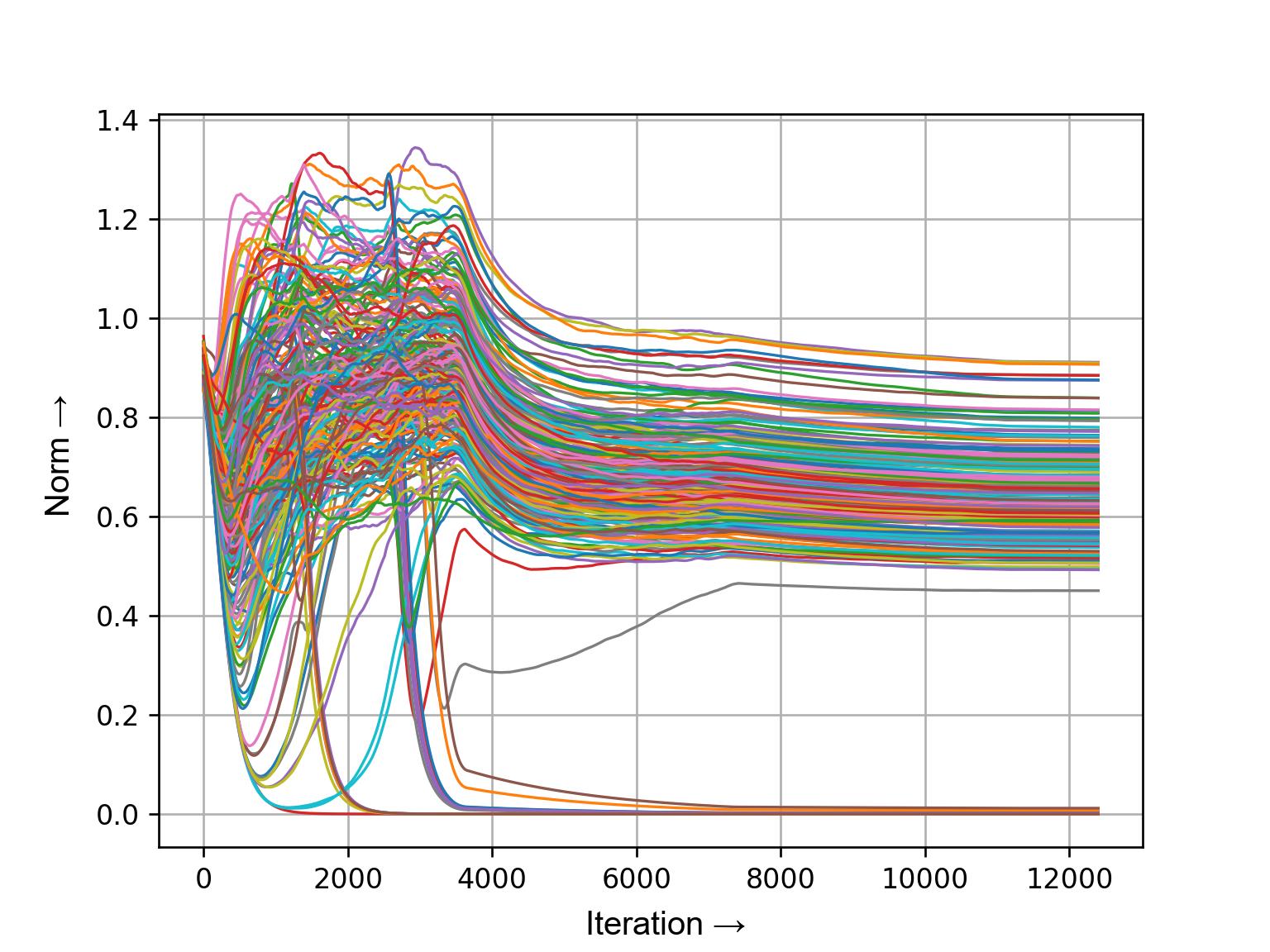}
\caption{Norm of filters of convolutional layer 11 (without PGT)}
\end{subfigure}
\begin{subfigure}[t]{0.32\textwidth}
\includegraphics[width=\textwidth]{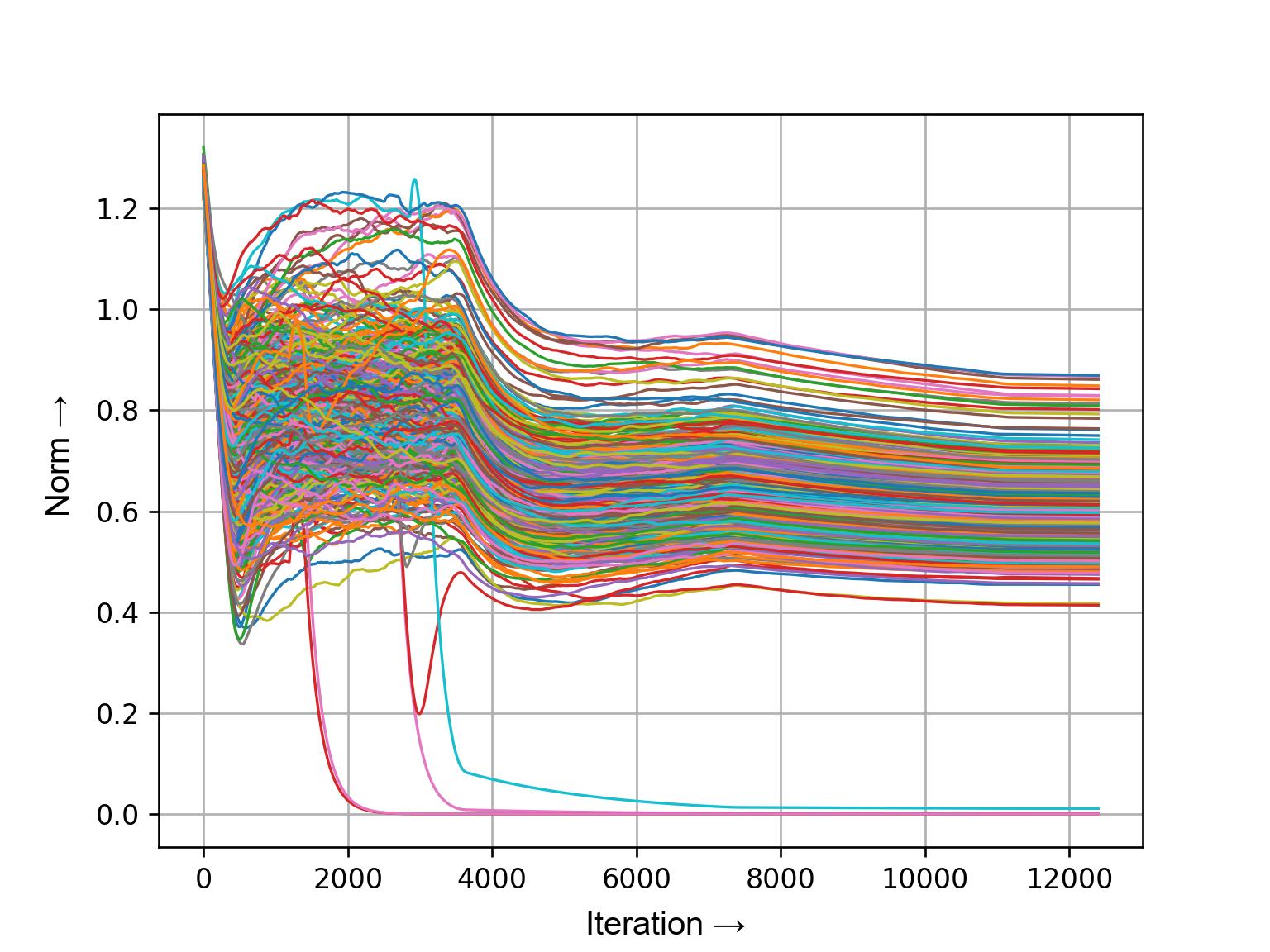}
\caption{Norm of filters of the final convolutional layer (without PGT)}
\end{subfigure}
\begin{subfigure}[t]{0.32\textwidth}
\includegraphics[width=\textwidth]{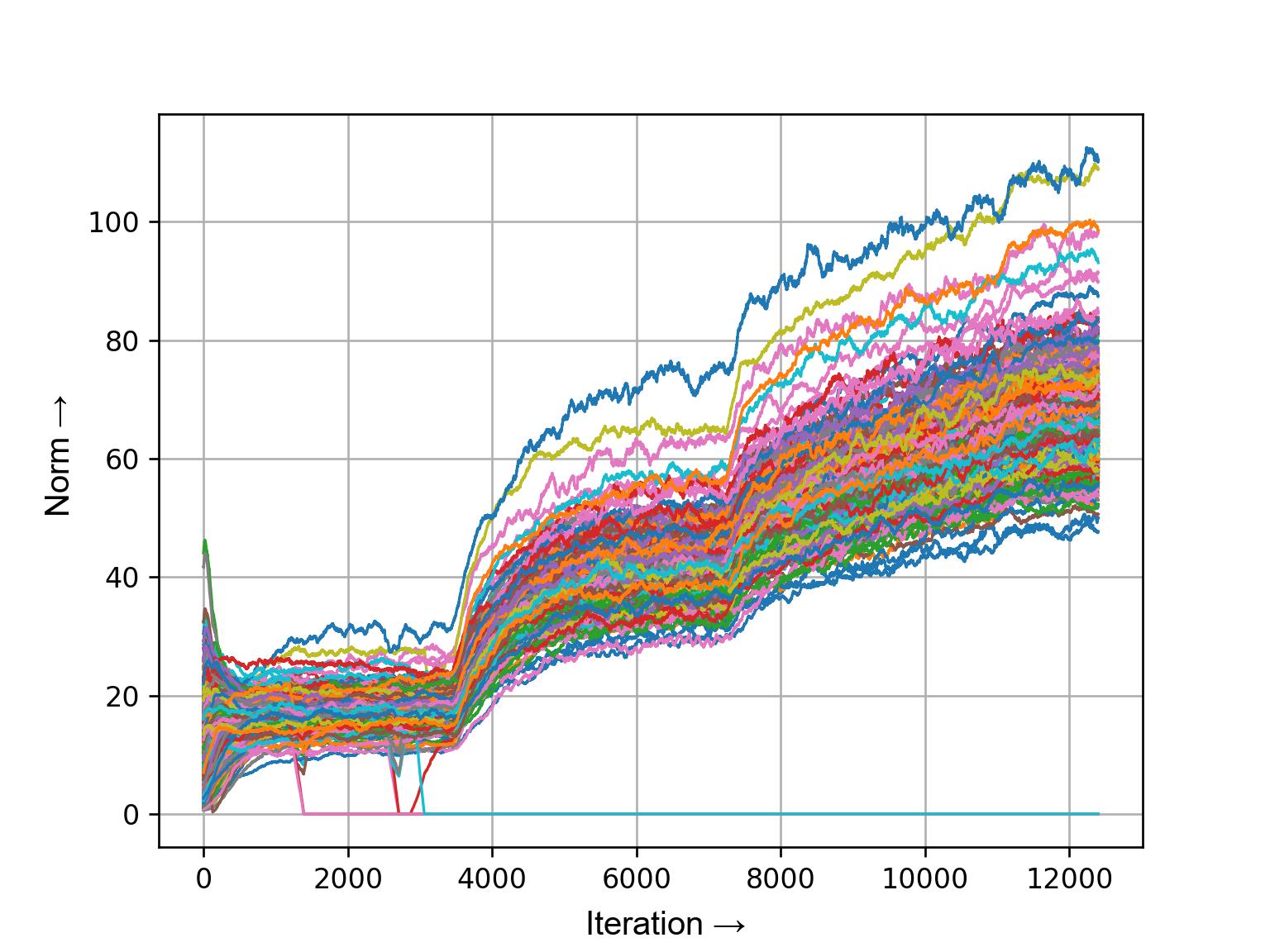}
\caption{Norm of final layer's output features (without PGT)}
\end{subfigure}

\begin{subfigure}[t]{0.32\textwidth}
\includegraphics[width=\textwidth]{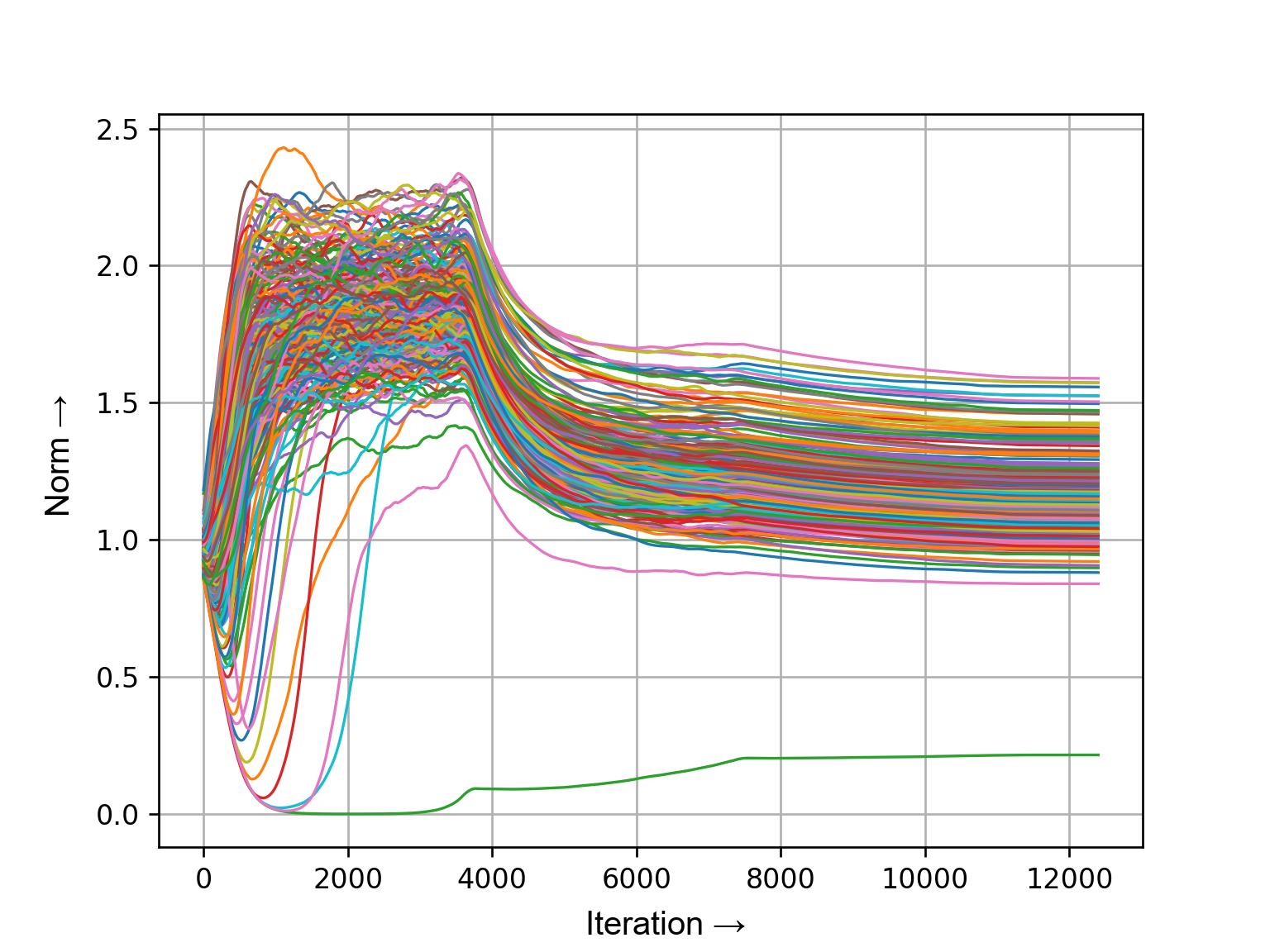}
\caption{Norm of filters of convolutional layer 11 (with PGT)}
\end{subfigure}
\begin{subfigure}[t]{0.32\textwidth}
\includegraphics[width=\textwidth]{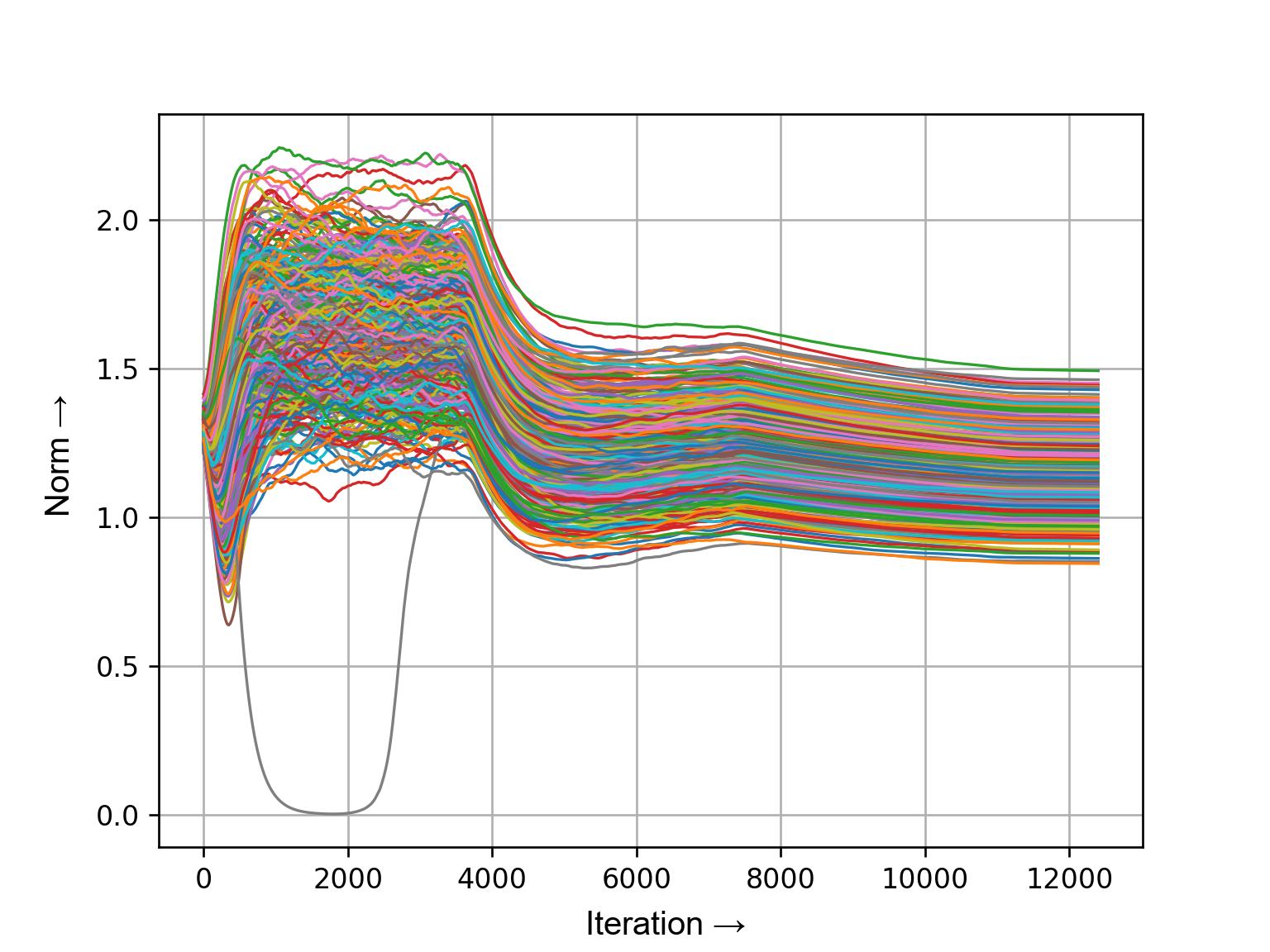}
\caption{Norm of filters of the final convolutional layer (with PGT)}
\end{subfigure}
\begin{subfigure}[t]{0.32\textwidth}
\includegraphics[width=\textwidth]{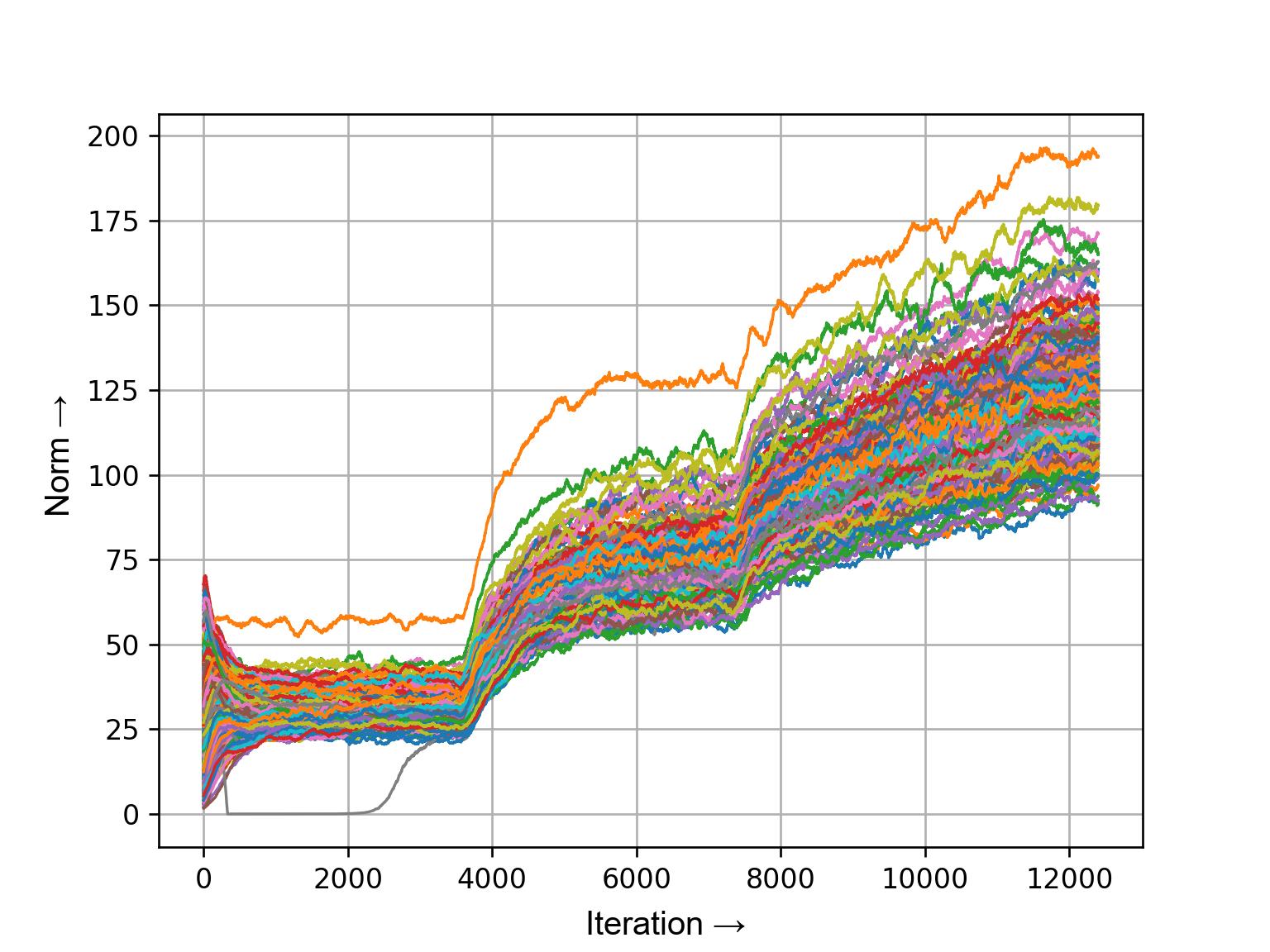}
\caption{Norm of final layer's output features (with PGT)}
\end{subfigure}
\captionsetup{font=normalsize}
\caption{ Experiment demonstrating the Zeroing Out phenomena, as observed in different
filters and feature vectors. PGT greatly reduces the zeroing out of filters and promote
better learning. }
\label{fig:norm_plots}
\end{figure}

We concentrate on the following metric: the per-filter L2-norm of each layer's weight
tensor. Throughout the training process, we monitor variations in the per-filter weight
norm. Because a full plot of each layer's data would use an inordinate amount of space,
we focus on a few critical layers' plots. The supplementary section contains full plots
for all layer for various settings. The norm of the weights of the $11^\textit{th}$
convolutional layer is shown in Fig. \ref{fig:norm_plots}(a). Layer $11$ includes $256$
filters. Each colour on the figure represents the iteration-wise evolution of a
particular filter throughout the course of training. The critical point to note is that
some filters achieve a norm of zero during training. We refer to this event as `Zeroing
Out', and occurs when a channel of a weight tensor gets fully filled with zeros. If all
the coefficients of the filter becomes zero, only then the L2 norm of that filter
becomes zero. If we observe zero filter norm for a filter, then that filter does not
contribute at all to determine the input-output relationship of a dataset, as the
feature tensor it produces is also an all-zero tensor. According to our findings, when a
filter tensor becomes zeroed out, it does not recover with further training. We observe
both effects \textbf{1)} zeroed out filter \textbf{2)} zeroed out feature.

Similar to the plot \ref{fig:norm_plots}(a), we see the plot of the final convolutional
layer in Fig. \ref{fig:norm_plots}(b), where we observe a number of filters / units
completely dropping out towards the end of training leading to underutilization of the
network's parameters. In Fig. \ref{fig:norm_plots}(c), we see the norm of the feature
vector at the output side of the global average pooling layer of ResNet-18. The feature
vector here directly interfaces with the fully connected layer immediately after. Any
zeroed out regions in this tensor directly leads to permanent information loss, as it
does not contribute to the learning of decision boundaries in the upcoming fully
connected layer, and indeed there exists a few zeroed out features in the baseline
model, which we discuss next. The impact of having a zeroed out filter or a feature
might have a variety of consequences for training and optimization. If a zeroed out
filter is present in a convolutional layer, there are two possible outcomes. If the
layer is immediately followed by a fully connected regression layer, there is an abrupt
loss of information. For the reason that a zeroed out filter always generates zeroed out
features regardless of the input image, when this tensor is fed to a fully connected
layer, it does not communicate any information that is useful for classification. In the
event that the zeroed out filter has a skip connection linked across it, this may be
circumvented as long as there are no overlapping zeroed out channels in the two feature
maps. On the other hand, if a feature tensor has a zeroed out channel at the input of a
layer, the layer creates a zeroed out feature channel at the output, regardless of the
weights. However, if there is a zeroed out channel in the feature map just before the
fully connected layer, then this results in a permanent loss of information as it does
not contribute to the learning of decision boundaries. ResNet-18 suffers from the same
kind of information loss, as the final feature tensor (which is generated after the
global average pooling layer) has multiple channels that are zeroed out [Fig.
\ref{fig:norm_plots}(c)]. This information collapse phenomenon occurs when a channel
that has been zeroed out stays zeroed out for all images in the dataset.

\begin{figure}[t]
\centering
\includegraphics[width=0.45\columnwidth]{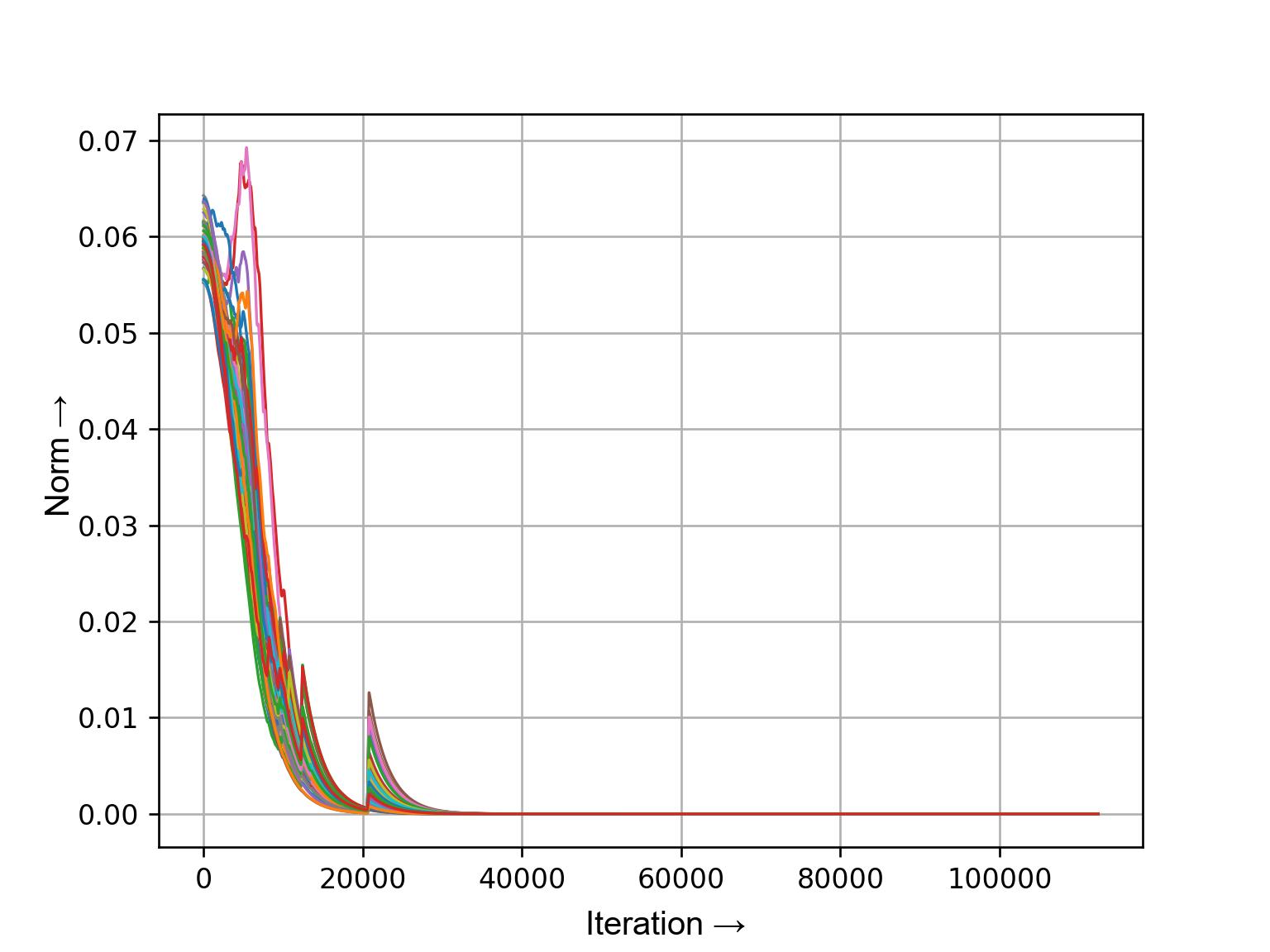}
\caption{ Per filter norm vs. iteration plot of layer 5 of an unnormalized ResNet-18,
trained with a very high batch size of 1024. The weight matrix of the entire layer
zeroes out during training. }
\label{fig:high_bs}
\end{figure}

The impact of a feature tensor or weight tensor being zeroed out is seen to be
particularly common in training procedures involving larger batch sizes in networks
without normalization layers. As a result, the network's parameters are underutilized
since some of the filters are not involved in extracting any useful information that is
important for performing classification. When working with large batch sizes, it is
possible that an entire layer zeroes out [Fig. \ref{fig:high_bs}]. In a network without
skip connections, this can cause a full collapse in training. For example, if a weight
tensor completely zeroes out, and it always produces a zero feature tensor, then the
input feature tensor to the next convolutional layer is a zero tensor and thus the
output of the next layer is always a zero tensor irrespective of its weights, and this
continues until the end of the neural network pipeline leading to zero logits. Also,
zeroed out weights tensors lead to zeroed out gradients hence stopping training for all
subsequent iterations leading to a collapse in training. Although residual networks
bypass this difficulty via the use of the skip connection, it does not solve the layer's
non-participation in the overall objective of classification. It is possible to solve
this issue by using methods such as gradient clipping (GC), adaptive gradient clipping
(AGC), and PowerGrad Transform (PGT), all of which regulate the amount of gradient that
flows at various junctions of the network. GC and AGC modify the gradients at each and
every filter of every layer. PGT performs this operation at only one junction, which is
at the softmax layer.

\begin{figure}[t]
\centering
\captionsetup{font=scriptsize}

\begin{subfigure}[t]{0.32\textwidth}
\includegraphics[width=\textwidth]{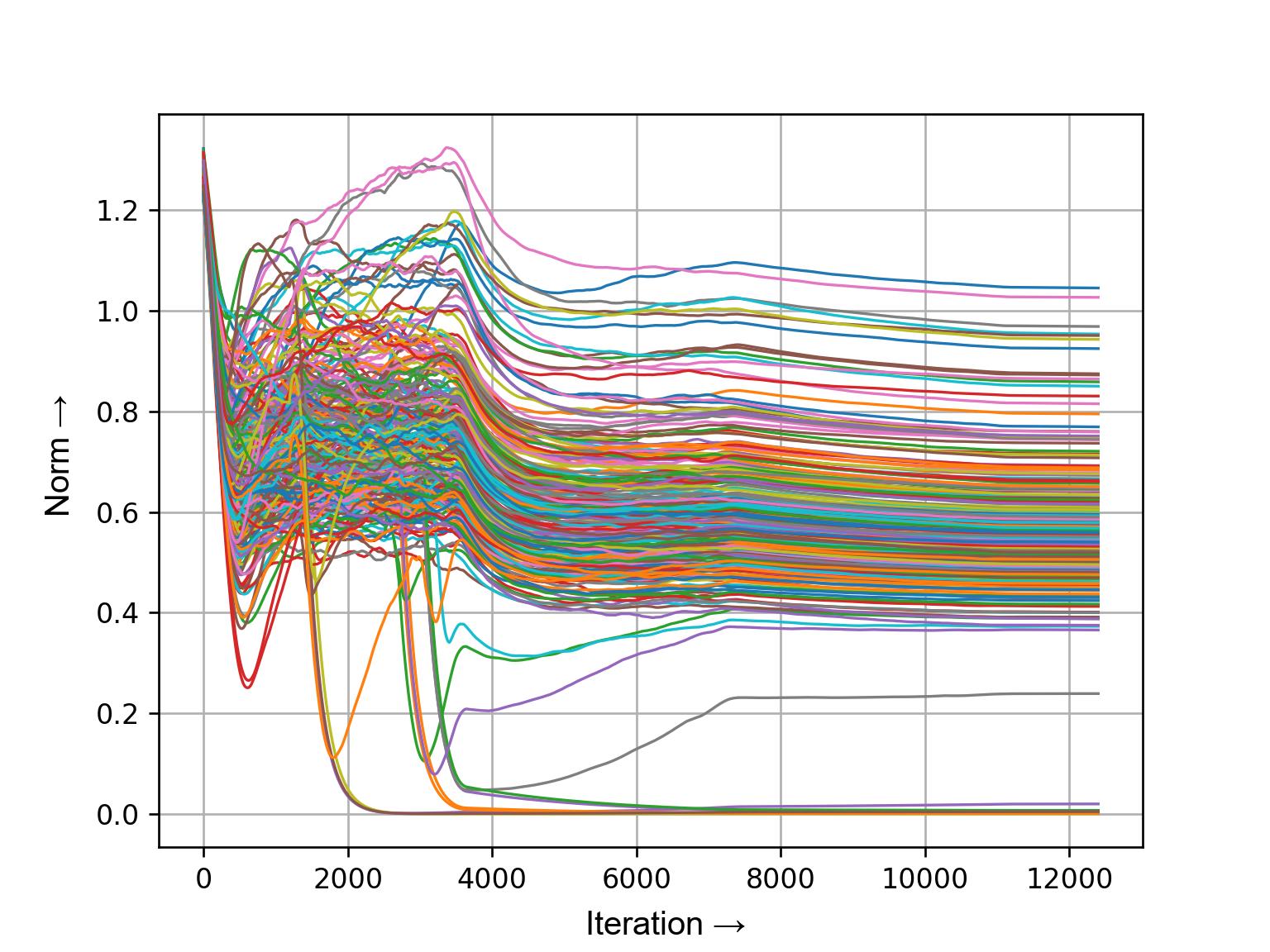}
\caption{Norm of filters of layer 15 (without AGC/PGT)}
\end{subfigure}
\begin{subfigure}[t]{0.32\textwidth}
\includegraphics[width=\textwidth]{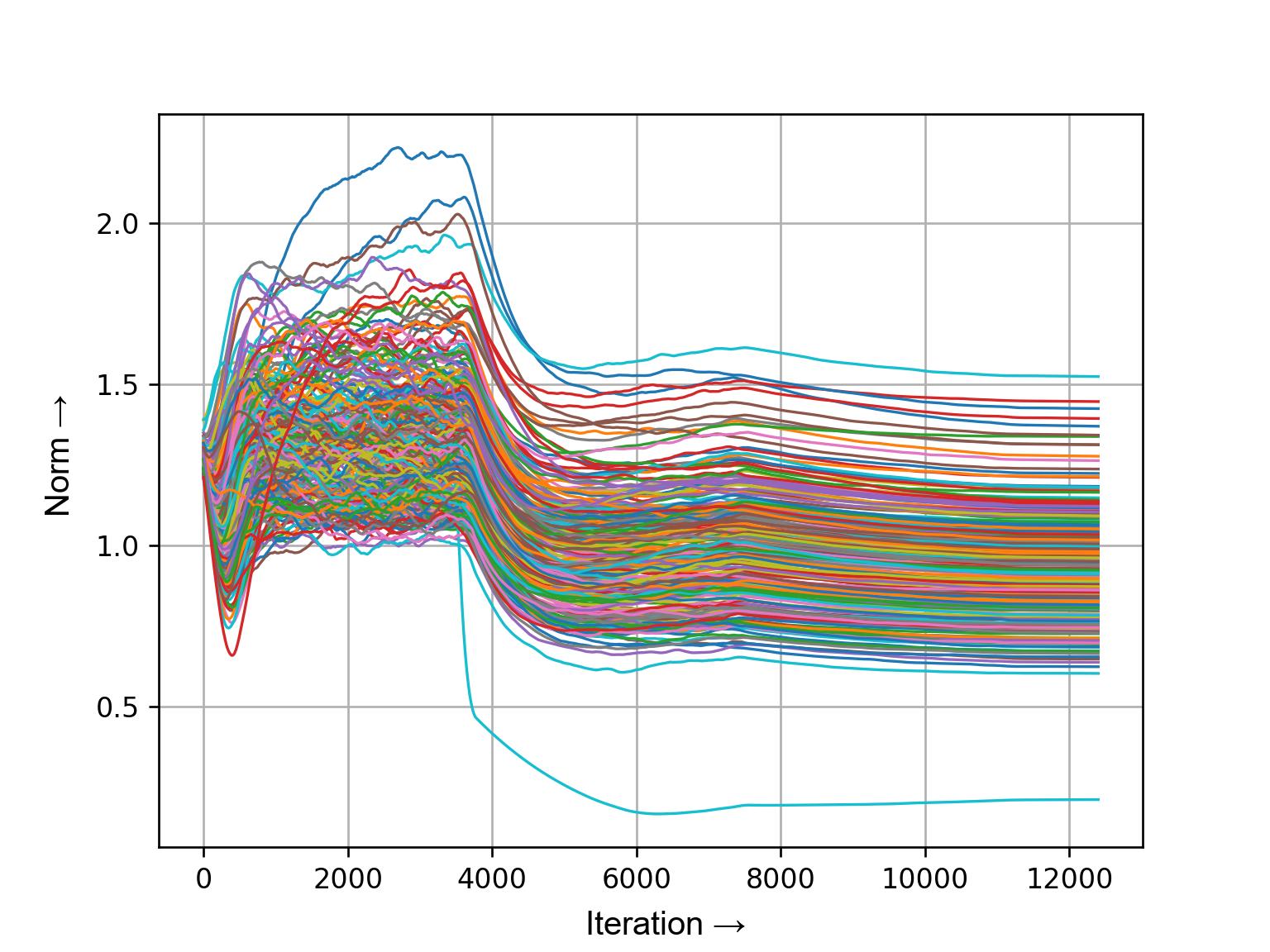}
\caption{Norm of filters of layer 15 (with PGT)}
\end{subfigure}
\begin{subfigure}[t]{0.32\textwidth}
\includegraphics[width=\textwidth]{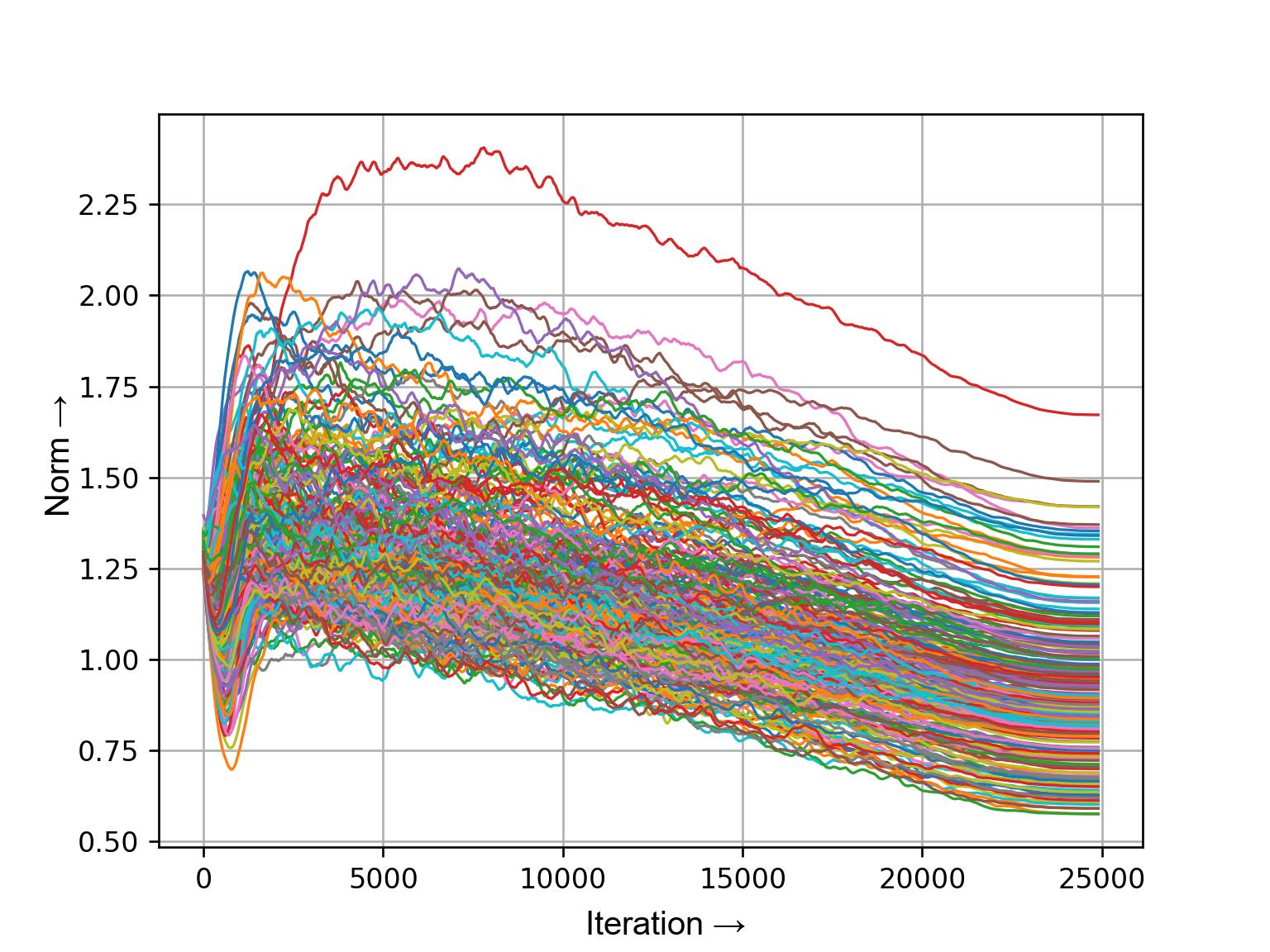}
\caption{Norm of filters of layer 15 (with AGC+PGT)}
\end{subfigure}

\begin{subfigure}[t]{0.32\textwidth}
\includegraphics[width=\textwidth]{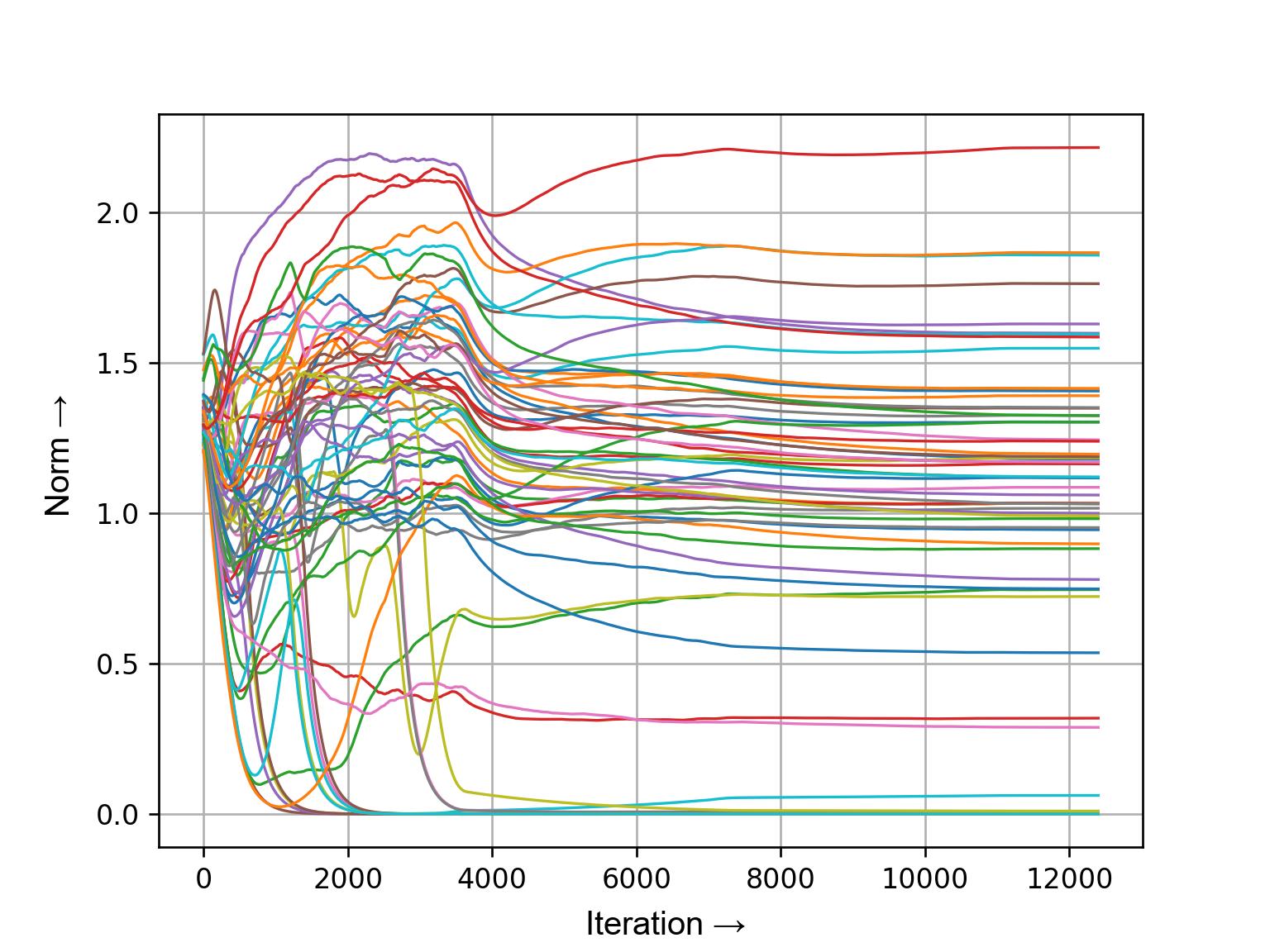}
\caption{Norm of filters of layer 2 (without AGC/PGT)}
\end{subfigure}
\begin{subfigure}[t]{0.32\textwidth}
\includegraphics[width=\textwidth]{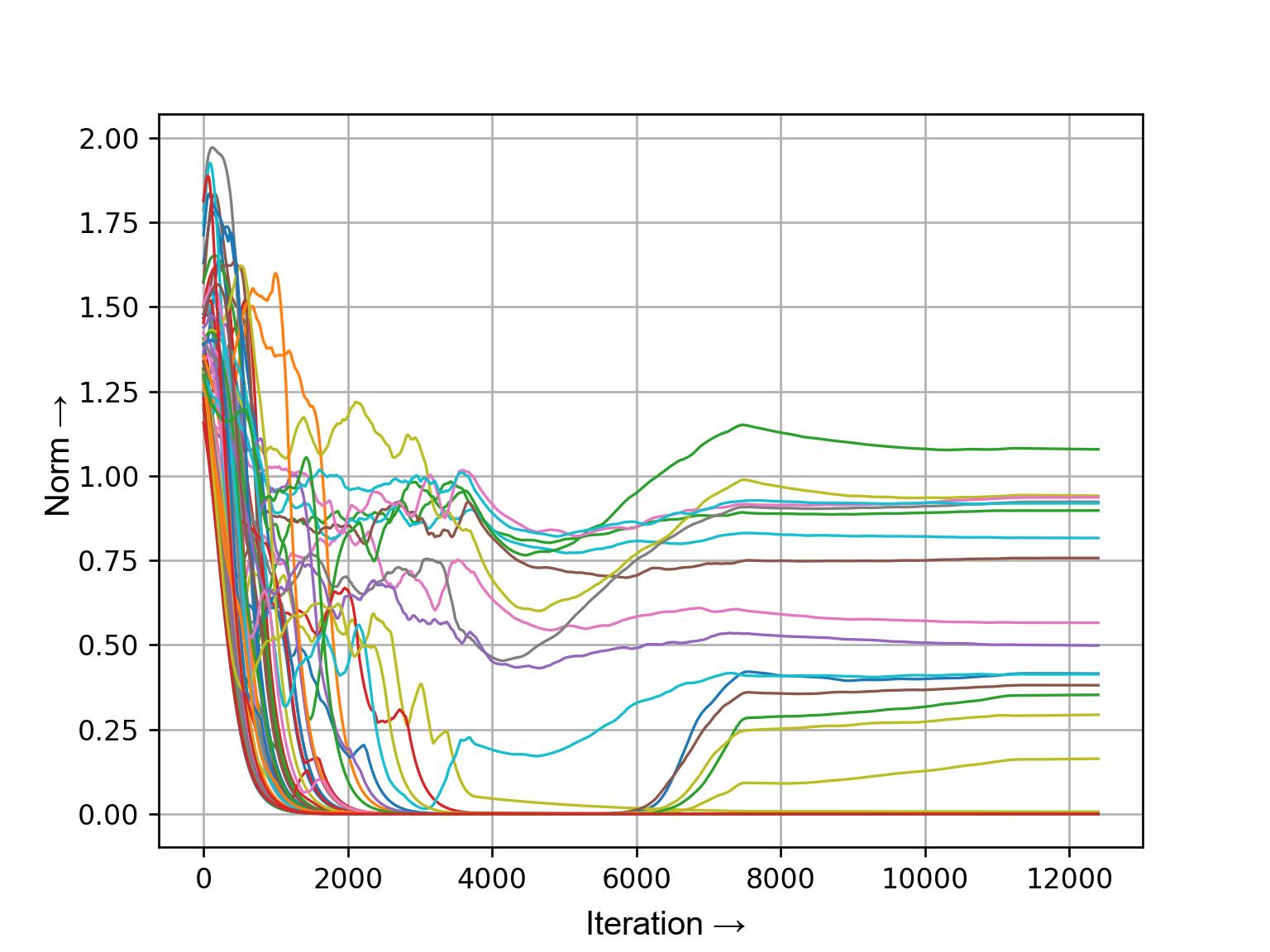}
\caption{Norm of filters of layer 2 (with PGT)}
\end{subfigure}
\begin{subfigure}[t]{0.32\textwidth}
\includegraphics[width=\textwidth]{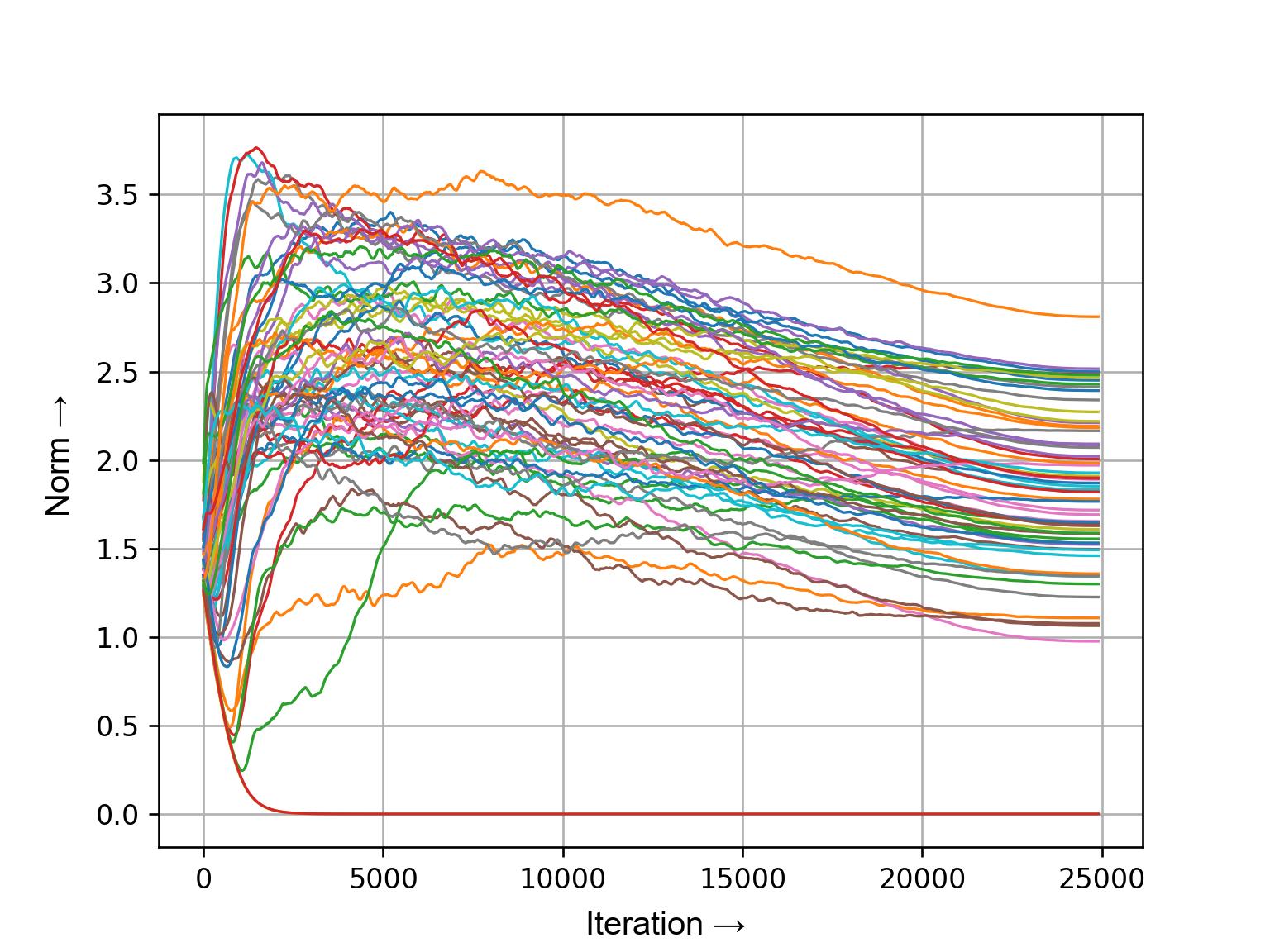}
\caption{Norm of filters of layer 2 (with AGC+PGT)}
\end{subfigure}
\captionsetup{font=normalsize}
\caption{ Experiment demonstrating the efficacy of PGT and AGC+PGT, as observed in
different layers. }
\label{fig:agc_pgt}
\end{figure}

In a similar fashion to the study in section \ref{sec:Empi}, we examine the norm
profiles of features and weights in the PGT and AGC. The supplementary section includes
all the detailed graphs of all layer. In Fig. \ref{fig:norm_plots}(d), we show the
per-filter norm plot of the $11^\textit{th}$ convolutional layer as it progresses
through a PGT enabled training session. We observe that the number of zeroed out filters
has considerably reduced. We observe similar benefits in plot of Fig.
\ref{fig:norm_plots}(e), which is the plot of the per-filter norm of the final
convolutional layer. Coming to benefits in feature norms, we observe in plot of Fig.
\ref{fig:norm_plots}(f), the final feature tensor obtained with PGT enabled training do
not contain any zeroed out regions, leading us to assert that information loss is
heavily mitigated as the features pass on from the feature extracting convolutional
layers to the fully connected stage for regression.

AGC is excellent at eliminating the issue of filter and feature zeroing out. However,
AGC recovers a lower performance benefit than PGT. In the instance of PGT, we see that
it also mitigates the zeroing out phenomenon in most layers. In Fig.
\ref{fig:agc_pgt}(a), we see a plot of the $15^\textit{th}$ layer of the non-BN
ResNet-18 trained without PGT. As we can see that there has been quite a few filter
dropouts in the baseline. PGT recovers most of the filters [Fig. \ref{fig:agc_pgt}(b)].
In some layers however, the performance of PGT is slightly weaker than the baseline. The
$2^\textit{nd}$ convolutional layer has more zeroed out filters in PGT [Fig.
\ref{fig:agc_pgt}(e)] than in the baseline [Fig. \ref{fig:agc_pgt}(d)]. Despite that,
PGT achieves a higher overall accuracy (Table \ref{tab:wobn_table}). This necessitates
the simultaneous use of AGC and PGT, for the non-BN ResNet-18.

\begin{table}[!t]
\centering
\caption{ Results for non-normalized ResNets on ImageNet-1K. Each experiment is
conducted with a 100 epoch budget using the cosine decay scheduler. Best training and
test accuracies are highlighted in red and blue respectively. Top differences in
training and test accuracies are marked in yellow. The baselines in each segment is
taken as the reference point for all `Diff' accuracies in that segment. The first three
segments highlight comparison for experiments where the methods are applied
individually, while the last segment denotes performance metrics for combinations of
different methods (GC + PGT, AGC + PGT). We see that at any batch size, with PGT, the
performance of non-normalized version of ResNet-18 improves significantly over the
baseline owing to the improvements in training as observed in Fig. \ref{fig:agc_pgt}. }
\label{tab:wobn_table}
\scalebox{0.65}{
\begin{tabular}{cccccccc}
\multirow{2}{*}{\textbf{Model}} & \multirow{2}{*}{\textbf{Batch Size}} &
\multirow{2}{*}{\textbf{Method}} & \textbf{PowerGrad} & \textbf{\hspace{-0.65cm} Train}
& \textbf{Train} & \textbf{\hspace{-0.65cm} Test} & \textbf{Test} \\
& & & \textbf{Transform ($\alpha$)} & \textbf{Accuracy (\%)} & \textbf{Diff (\%)} &
\textbf{Accuracy (\%)} & \textbf{Diff (\%)} \\
\midrule
ResNet-18 (non-BN) & 1024 & Baseline & - & 66.27 & - & 64.816 & - \\
& 1024 & PGT & 0.92 & \textcolor{red}{\textbf{66.62}} &
\textcolor{olive}{\textbf{+0.35}} & \textcolor{blue}{\textbf{65.498}} &
\textcolor{olive}{\textbf{+0.682}} \\
\cmidrule{2-8}
& 512 & Baseline & - & 68.02 & - & 66.552 & - \\
& 512 & PGT & 0.25 & \textcolor{red}{\textbf{69.5}} & \textcolor{olive}{\textbf{+1.48}}
& \textcolor{blue}{\textbf{67.236}} & \textcolor{olive}{\textbf{+0.684}} \\
\cmidrule{2-8}
& 256 & Baseline & - & 68.86 & - & 66.796 & - \\
& 256 & GC & - & 69.04 & +0.18 & 67.064 & +0.268 \\
& 256 & AGC & - & 69.06 & +0.2 & 67.298 & +0.502 \\
& 256 & PGT & 0.25 & \textcolor{red}{\textbf{69.97}} & \textcolor{olive}{\textbf{+1.11}}
& \textcolor{blue}{\textbf{67.814}} & \textcolor{olive}{\textbf{+1.018}} \\
\cmidrule{2-8}
& 256 & Baseline & - & 68.86 & - & 66.796 & - \\
& 256 & GC+PGT & 0.25 & 68.67 & -0.19 & 67.088 & +0.292 \\
& 256 & AGC+PGT & 0.25 & \textcolor{red}{\textbf{70.92}} &
\textcolor{olive}{\textbf{+2.06}} & \textcolor{blue}{\textbf{68.856}} &
\textcolor{olive}{\textbf{+2.06}} \\
\end{tabular}}
\end{table}

We provide performance metrics on the test set of ImageNet-1K for different batch sizes
for the non-BN ResNet-18 and also experiment with different methods (GC, AGC, PGT along
with their combinations). Table \ref{tab:wobn_table} shows the findings. To begin, we
use a batch size of 1024. The baseline performance for this batch size is drastically
inferior to baselines for other batch sizes. PGT helps regain some of the lost
performance by $0.682\%$ ($65.498\%$ vs. $64.816\%$). This illustrates that in
unnormalized networks, accuracy suffers significantly when batch sizes are large.
Additionally, the training process is often unstable at this batch size, resulting in
filters and sometimes entire layers frequently zeroing out (as seen in section
\ref{sec:Empi}). At a batch size of 512, invoking PGT improves the training accuracy
baseline by $1.48\%$ and the test accuracy baseline by $0.684\%$, while at a batch size
of 256, the improvement in training and test accuracies are $1.11\%$ and $1.018\%$
respectively. In comparison, the test accuracy improvements obtained by GC and AGC at
batch size of 256 is much less, at $0.27\%$ and $0.5\%$, respectively.

On the training accuracy front, since we get a significant boost ($1.48\%$ at batch size
of 512 and $1.11\%$ at batch size of 256), it leads us to infer that when PowerGrad
Transform is used, the network fits the training dataset more tightly and the
convergence optima is significantly superior. We also notice that the training versus
testing gap increases with PGT as compared to the gap at baseline even though both
training and test accuracies improve. Therefore it allows us to conclude that the
improvements in training and test accuracies are obtained not through regularization
like it does in methods such as label smoothing; rather the improvements are obtained as
PGT enables the network to utilize its learning capacity better and arrive at a higher
performing optima at convergence, something which is inaccessible through traditional
gradient based training.

We also experiment with combinations of GC/AGC and PGT. The advantage of combining GC
with PGT is not very significant. When AGC and PGT are combined, we see a tremendous
increase in test accuracy of over $2.06\%$ over the baseline. As seen in Fig.
\ref{fig:agc_pgt}(c) and \ref{fig:agc_pgt}(f), activating AGC and PGT concurrently
results in no filter dropouts in all layers while also improving the accuracy by a
significant margin. Additionally, we provide comprehensive graphs of the norm profiles
of the AGC + PGT method for all layers in the supplementary section.

\subsection{Effect on Loss, Logits and other metrics}
\label{sec:Effe}

\begin{figure}[t]
\centering
\begin{subfigure}{0.33\textwidth}
\centering
\includegraphics[width=0.98\textwidth]{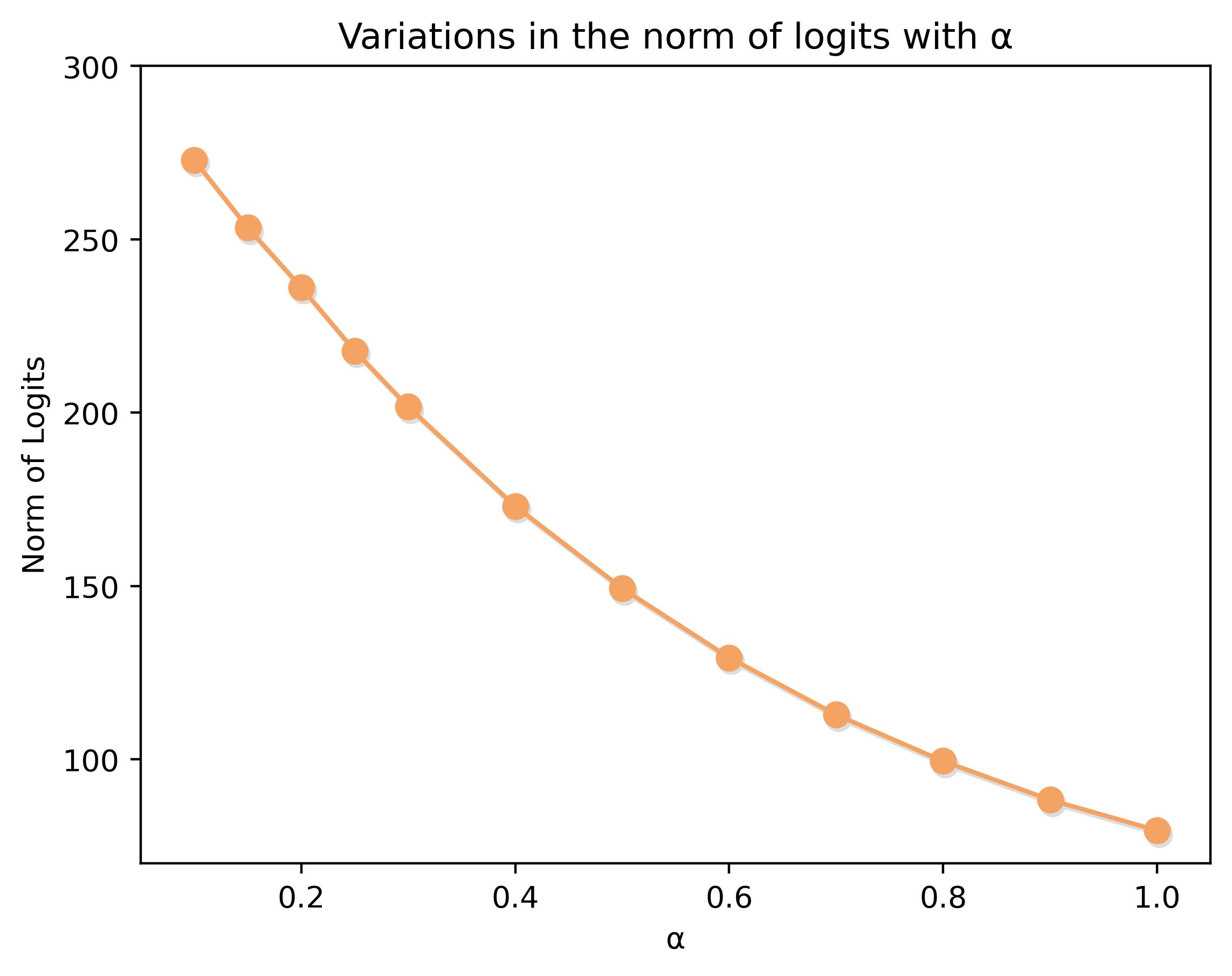}
\caption{\\Logit vs. $\alpha$}
\end{subfigure}%
\begin{subfigure}{0.33\textwidth}
\centering
\includegraphics[width=0.98\textwidth]{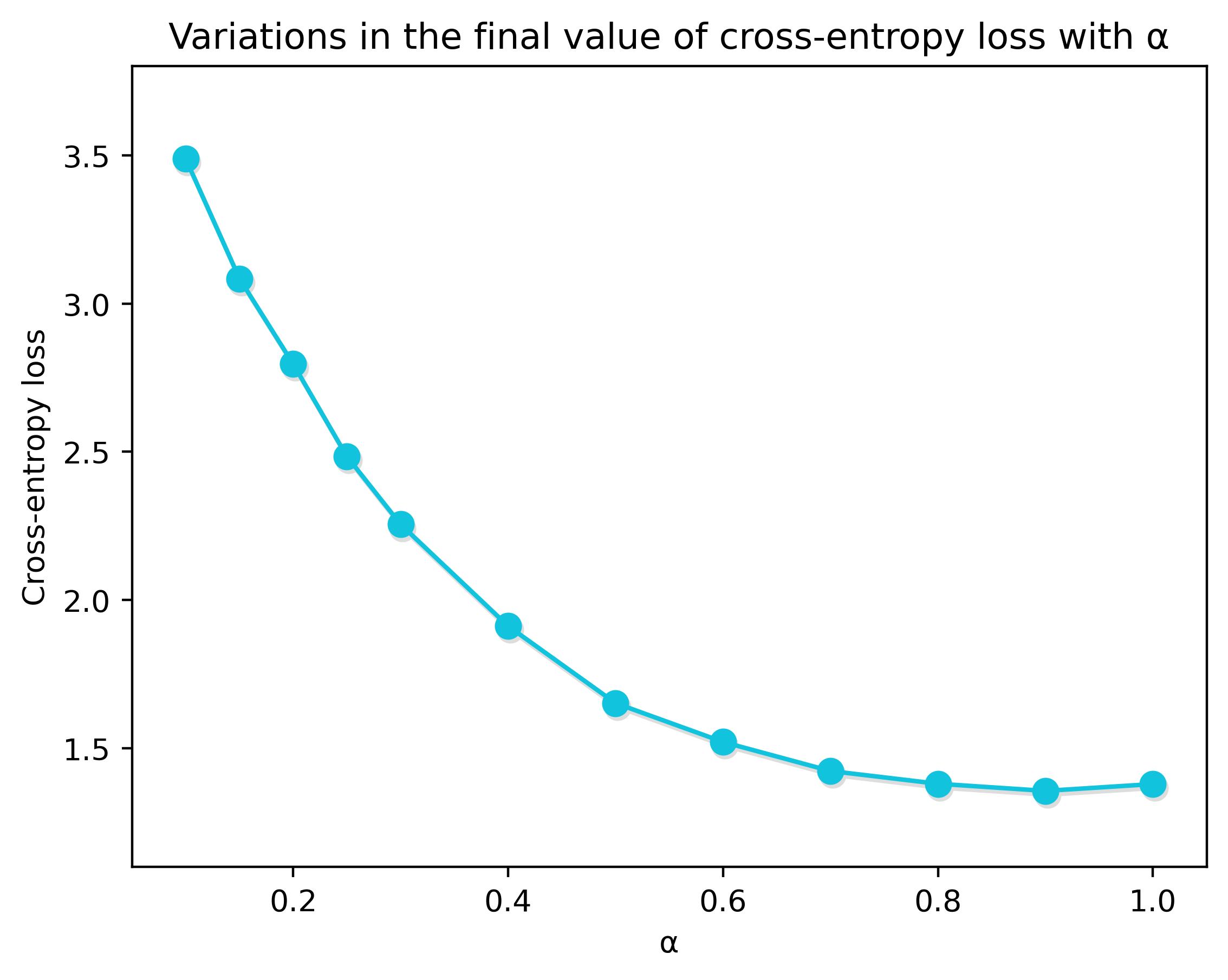}
\caption{\\Cross-entropy loss vs. $\alpha$}
\end{subfigure}%
\begin{subfigure}{0.33\textwidth}
\centering
\includegraphics[width=0.98\textwidth]{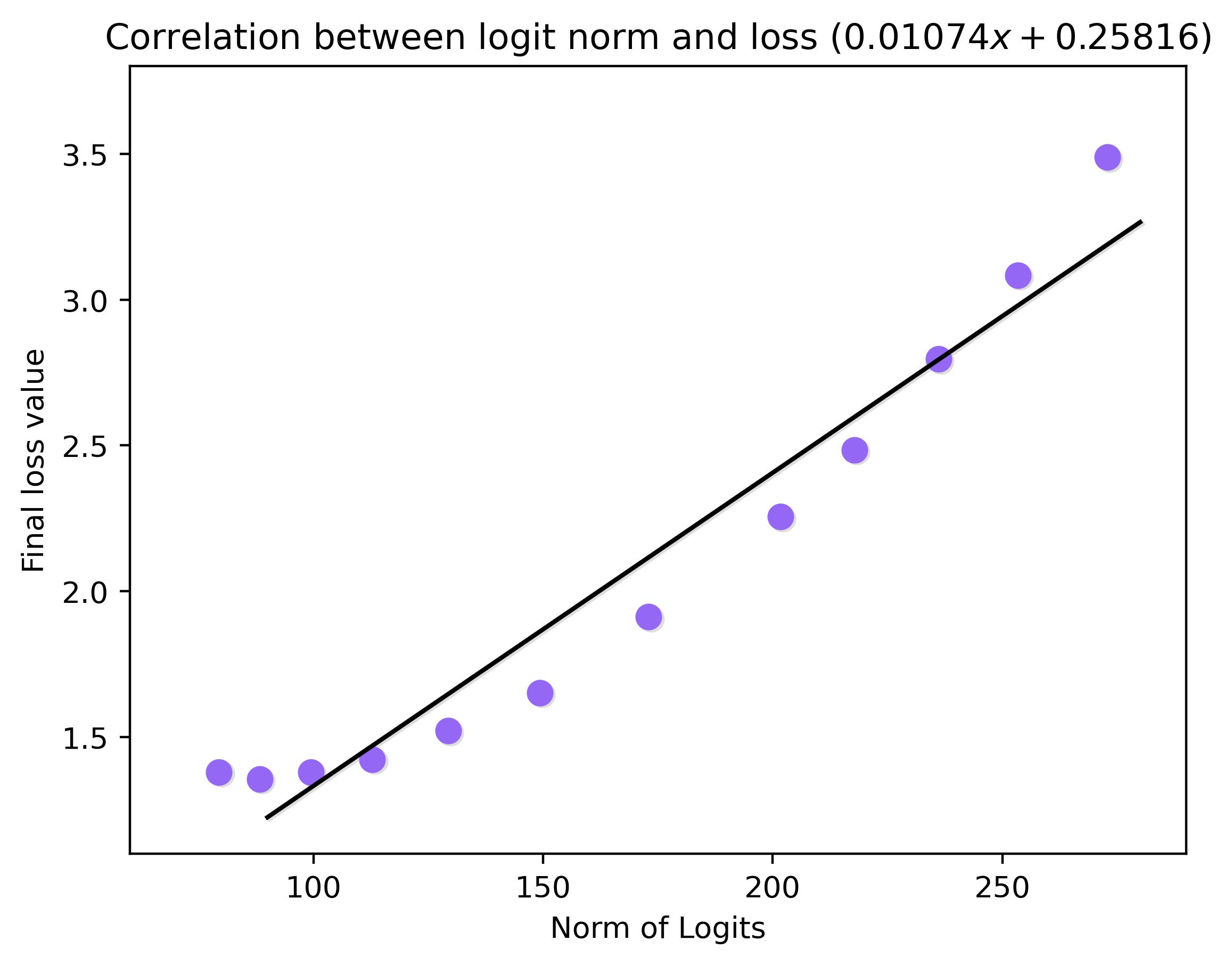}
\caption{\\Logit-Loss correlation}
\end{subfigure}
\caption{ Plots of various statistical measures: \textbf{(a)} Variations in the logit
norm vs. $\alpha$. The logit norm is calculated per image over the test set of
ImageNet-1K (at the linear layer of the ResNet-18 architecture) and then averaged.
\textbf{(b)} Variations in the final loss values obtained for different $\alpha$
settings. \textbf{(c)} Correlation between loss and logits. Regression line equation
:$(0.01074 x + 0.25816)$. }
\label{fig:stats}
\end{figure}

As seen in Fig. \ref{fig:stats}(a), the logit norm increases as $\alpha$ falls from $1$
to $0$. In the initial training phase, PGT causes the gradient in both the valid and
wrong classes to rise as the network misclassifies majority of the examples. Due to the
exponential nature of the softmax function, the predicted probability distribution may
frequently be close to a delta distribution, even in instances of misclassifications.
PGT reshapes the predicted probability distribution. It redistributes probability from
the confident classes to the less confident ones. Gradient rises in all classes. We also
see that the final value of the loss rises as $\alpha$ drops, and that the logit norm
and the final value of the loss are linearly correlated Fig. \ref{fig:stats}(c). What is
surprising is that: \textit{it is possible to achieve higher accuracies even though the
loss values are larger during convergence}. In our studies with ResNet-50, for instance,
if PGT is not invoked, the corresponding test accuracy is $76.56\%$ (Table
\ref{tab:imagenet_table}) with a cross-entropy loss value of $0.852$. However, if PGT is
activated with $\alpha=0.05$, the corresponding test accuracy is $77.216\%$ (Table
\ref{tab:imagenet_table}) with a loss value of $2.360$. This is markedly different from
the coupled gradient descent based training procedures. PGT enables the network to
arrive at such optima where the loss values are high, but both training and test
perfomance is better. From these plots of logits and losses we find that with the
inclusion of PGT in the training process, the model can have access to such regions of
the loss landscape that is otherwise inaccessible to traditional gradient based training
procedures.

\section{Ablation Study}
\label{sec:Abla}

\begin{table}[!t]
\centering
\caption{ Ablation study for ResNet-50 on ImageNet-1K. LS column denotes whether label
smoothing has been applied. A smoothing hyperparameter value of $0.1$ has been used. The
hyperparameter column that controls the amount of PGT gradient modification injected at
softmax, Eq. \ref{transformed_probabilities}. }
\label{tab:ablation_table}
\scalebox{0.7}{
\begin{tabular}{cccccccc}
\multirow{2}{*}{\textbf{\#Row}} & \multirow{2}{*}{\textbf{Model}} &
\multirow{2}{*}{\textbf{Scheduler}} & \textbf{Label} & \textbf{PowerGrad} &
\textbf{Train} & \textbf{Test} & \multirow{2}{*}{\textbf{Gap (\%)}} \\
& & & \textbf{Smoothing} & \textbf{Transform ($\alpha$)} & \textbf{Accuracy (\%)} &
\textbf{Accuracy (\%)} & \\
\midrule
1. & ResNet-50 & Step & \xmark & \xmark & 78.99 & 75.97 & 3.02 \\
2. & ResNet-50 & Step & \xmark & 0.3 & 79.56 & 76.494 & 3.066 \\
\cmidrule{2-8}
3. & ResNet-50 & Cosine & \xmark & \xmark & 79.18 & 76.56 & 2.62 \\
4. & ResNet-50 & Cosine & 0.1 & \xmark & 78.81 & 76.698 & 2.112 \\
\cmidrule{2-8}
5. & ResNet-50 & Cosine & \xmark & 0.3 & 79.43 & 76.886 & 2.544 \\
6. & ResNet-50 & Cosine & 0.1 & 0.3 & 78.47 & 76.968 & 1.502 \\
\cmidrule{2-8}
7. & ResNet-50 & Cosine & \xmark & 0.05 & 79.68 & 77.216 & 2.464 \\
8. & ResNet-50 & Cosine & 0.1 & 0.05 & 77.69 & 76.39 & 1.3 \\
\end{tabular}}
\end{table}

We conduct an ablation study to investigate the effects of PowerGrad Transform for
different values of the hyperparameter ($\alpha$). Grid plots are made available in the
supplementary section. We have previously demonstrated from the experiments depicted in
section \ref{sec:Expe}, that PowerGrad Transform causes the network to arrive at a
better optima and fit the training dataset better than the baseline. In this ablation
study, we use the ResNet-50 architecture and we combine our proposed method with other
regularization techniques such as label smoothing and report our findings in Table
\ref{tab:ablation_table}. We also investigate the two optimal peaks of the
hyperparameter ($\alpha$) obtained from ResNet-50's grid search ($\alpha=0.05$ and
$\alpha=0.3$). First, we examine the effect of PGT on the step scheduler baseline in
order to later compare it to the cosine scheduler baseline. \textbf{Row-1)} We begin
with the step scheduler baseline ($75.97\%$). \textbf{Row-2)} PGT improves upon the step
scheduler baseline (test set) by a substantial margin with $0.524\%$ ($76.494\%$ as
opposed to $75.97\%$). \textbf{Row-3)} Introducing the cosine scheduler yields a
$0.59\%$ improvement ($76.56\%$ vs. $75.97\%$) over the step scheduler. \textbf{Row-4)}
After introducing label smoothing, the test accuracy relative to the cosine scheduler
baseline increases by only $0.138\%$ (from $76.56\%$ to $76.698\%$). \textbf{Row-5)}
However, introducing PGT with $\alpha=0.3$ alone (without label smoothing) improves the
cosine scheduler baseline by $0.326\%$ ($76.886\%$ vs. $76.56\%$). \textbf{Row-6)}
Combining PGT ($\alpha=0.3$) with label smoothing improves the performance on the test
set further by $0.408\%$ (from $76.56\%$ to $76.968\%$) and reduces the generalization
gap (from $2.54\%$ to $1.5\%$). However, the impact of combining PGT with label
smoothing can vary depending on the value of the hyperparameter ($\alpha$).
\textbf{Row-7)} With a PGT hyperparameter value of $\alpha=0.05$, we notice the greatest
performance improvement, $1.246\%$ over the step scheduler test baseline and $0.656\%$
over the cosine scheduler test baseline. \textbf{Row-8)} Adding label smoothing to PGT
($\alpha=0.05$) hurts performance even though it reduces the generalization gap.

\section{Discussions}
\label{sec:Disc}

We introduced PowerGrad Transform, which decouples the backward and forward passes of
neural network training and enables a significantly better fit to the dataset, as
measured by training and test accuracy metrics. We show the application of PowerGrad
Transform, a simple yet powerful technique for modifying the gradient flow behaviour. We
investigate experimentally the degenerate behavior of non-BN networks, which is
frequently observed during the standard training approach, especially with higher batch
sizes. With PGT, gradient behavior is enhanced and the likelihood of weights attaining
degenerate states is diminished. We provide a theoretical analysis of the PowerGrad
transformation. With the use of different network topologies and datasets, we are able
to show the potential of PGT and explore its impacts from an empirical standpoint. We
provide comprehensive results on a number of models (non-BN and BN ResNets, SE-ResNets)
using the ImageNet dataset. PGT helps the network to improve its learning capabilities
by locating a more optimum convergence point and simultaneously speeds up training. In
addition, we undertake an ablation research and compare its effects to those of
regularization methods such label smoothing.

\bibliographystyle{plain}
\bibliography{arxiv}

\begin{thebibliography}{10}

\bibitem{aguilar2020knowledge}
Gustavo Aguilar, Yuan Ling, Yu~Zhang, Benjamin Yao, Xing Fan, and Chenlei Guo.
\newblock Knowledge distillation from internal representations.
\newblock In {\em Proceedings of the AAAI Conference on Artificial
  Intelligence}, volume~34, pages 7350--7357, 2020.

\bibitem{aladago2021slot}
Maxwell~M Aladago and Lorenzo Torresani.
\newblock Slot machines: Discovering winning combinations of random weights in
  neural networks.
\newblock In {\em International Conference on Machine Learning}, pages
  163--174. PMLR, 2021.

\bibitem{DBLP:conf/nips/BaC14}
Jimmy Ba and Rich Caruana.
\newblock Do deep nets really need to be deep?
\newblock In Zoubin Ghahramani, Max Welling, Corinna Cortes, Neil~D. Lawrence,
  and Kilian~Q. Weinberger, editors, {\em Advances in Neural Information
  Processing Systems 27: Annual Conference on Neural Information Processing
  Systems 2014, December 8-13 2014, Montreal, Quebec, Canada}, pages
  2654--2662, 2014.

\bibitem{brock2021high}
Andrew Brock, Soham De, Samuel~L Smith, and Karen Simonyan.
\newblock High-performance large-scale image recognition without normalization.
\newblock {\em arXiv preprint arXiv:2102.06171}, 2021.

\bibitem{cho2019efficacy}
Jang~Hyun Cho and Bharath Hariharan.
\newblock On the efficacy of knowledge distillation.
\newblock In {\em Proceedings of the IEEE/CVF International Conference on
  Computer Vision}, pages 4794--4802, 2019.

\bibitem{chorowski2016towards}
Jan Chorowski and Navdeep Jaitly.
\newblock Towards better decoding and language model integration in sequence to
  sequence models.
\newblock {\em arXiv preprint arXiv:1612.02695}, 2016.

\bibitem{deng2009imagenet}
Jia Deng, Wei Dong, Richard Socher, Li-Jia Li, Kai Li, and Li~Fei-Fei.
\newblock Imagenet: A large-scale hierarchical image database.
\newblock In {\em 2009 IEEE conference on computer vision and pattern
  recognition}, pages 248--255. Ieee, 2009.

\bibitem{dubey2018pairwise}
Abhimanyu Dubey, Otkrist Gupta, Pei Guo, Ramesh Raskar, Ryan Farrell, and
  Nikhil Naik.
\newblock Pairwise confusion for fine-grained visual classification.
\newblock In {\em Proceedings of the European conference on computer vision
  (ECCV)}, pages 70--86, 2018.

\bibitem{duchi2011adaptive}
John Duchi, Elad Hazan, and Yoram Singer.
\newblock Adaptive subgradient methods for online learning and stochastic
  optimization.
\newblock {\em Journal of machine learning research}, 12(7), 2011.

\bibitem{furlanello2018born}
Tommaso Furlanello, Zachary Lipton, Michael Tschannen, Laurent Itti, and Anima
  Anandkumar.
\newblock Born again neural networks.
\newblock In {\em International Conference on Machine Learning}, pages
  1607--1616. PMLR, 2018.

\bibitem{ge2019distilling}
Shiming Ge, Zhao Luo, Chunhui Zhang, Yingying Hua, and Dacheng Tao.
\newblock Distilling channels for efficient deep tracking.
\newblock {\em IEEE Transactions on Image Processing}, 29:2610--2621, 2019.

\bibitem{ge2018low}
Shiming Ge, Shengwei Zhao, Chenyu Li, and Jia Li.
\newblock Low-resolution face recognition in the wild via selective knowledge
  distillation.
\newblock {\em IEEE Transactions on Image Processing}, 28(4):2051--2062, 2018.

\bibitem{he2016deep}
Kaiming He, Xiangyu Zhang, Shaoqing Ren, and Jian Sun.
\newblock Deep residual learning for image recognition.
\newblock In {\em Proceedings of the IEEE conference on computer vision and
  pattern recognition}, pages 770--778, 2016.

\bibitem{he2018determining}
Yu-Lin He, Xiao-Liang Zhang, Wei Ao, and Joshua~Zhexue Huang.
\newblock Determining the optimal temperature parameter for softmax function in
  reinforcement learning.
\newblock {\em Applied Soft Computing}, 70:80--85, 2018.

\bibitem{hinton2015distilling}
Geoffrey Hinton, Oriol Vinyals, and Jeff Dean.
\newblock Distilling the knowledge in a neural network.
\newblock {\em arXiv preprint arXiv:1503.02531}, 2015.

\bibitem{hu2018squeeze}
Jie Hu, Li~Shen, and Gang Sun.
\newblock Squeeze-and-excitation networks.
\newblock In {\em Proceedings of the IEEE conference on computer vision and
  pattern recognition}, pages 7132--7141, 2018.

\bibitem{lecun1988theoretical}
Yann LeCun, D~Touresky, G~Hinton, and T~Sejnowski.
\newblock A theoretical framework for back-propagation.
\newblock In {\em Proceedings of the 1988 connectionist models summer school},
  volume~1, pages 21--28, 1988.

\bibitem{lillicrap2020backpropagation}
Timothy~P Lillicrap, Adam Santoro, Luke Marris, Colin~J Akerman, and Geoffrey
  Hinton.
\newblock Backpropagation and the brain.
\newblock {\em Nature Reviews Neuroscience}, 21(6):335--346, 2020.

\bibitem{liu2017sphereface}
Weiyang Liu, Yandong Wen, Zhiding Yu, Ming Li, Bhiksha Raj, and Le~Song.
\newblock Sphereface: Deep hypersphere embedding for face recognition.
\newblock In {\em Proceedings of the IEEE conference on computer vision and
  pattern recognition}, pages 212--220, 2017.

\bibitem{liu2018improving}
Xuan Liu, Xiaoguang Wang, and Stan Matwin.
\newblock Improving the interpretability of deep neural networks with knowledge
  distillation.
\newblock In {\em 2018 IEEE International Conference on Data Mining Workshops
  (ICDMW)}, pages 905--912. IEEE, 2018.

\bibitem{loshchilov2016sgdr}
Ilya Loshchilov and Frank Hutter.
\newblock Sgdr: Stochastic gradient descent with warm restarts.
\newblock {\em arXiv preprint arXiv:1608.03983}, 2016.

\bibitem{merity2017regularizing}
Stephen Merity, Nitish~Shirish Keskar, and Richard Socher.
\newblock Regularizing and optimizing lstm language models.
\newblock {\em arXiv preprint arXiv:1708.02182}, 2017.

\bibitem{DBLP:conf/nips/MullerKH19}
Rafael M{\"{u}}ller, Simon Kornblith, and Geoffrey~E. Hinton.
\newblock When does label smoothing help?
\newblock In Hanna~M. Wallach, Hugo Larochelle, Alina Beygelzimer, Florence
  d'Alch{\'{e}}{-}Buc, Emily~B. Fox, and Roman Garnett, editors, {\em Advances
  in Neural Information Processing Systems 32: Annual Conference on Neural
  Information Processing Systems 2019, NeurIPS 2019, December 8-14, 2019,
  Vancouver, BC, Canada}, pages 4696--4705, 2019.

\bibitem{pascanu2012understanding}
Razvan Pascanu, Tomas Mikolov, and Yoshua Bengio.
\newblock Understanding the exploding gradient problem.
\newblock {\em CoRR, abs/1211.5063}, 2(417):1, 2012.

\bibitem{pascanu2013difficulty}
Razvan Pascanu, Tomas Mikolov, and Yoshua Bengio.
\newblock On the difficulty of training recurrent neural networks.
\newblock In {\em International conference on machine learning}, pages
  1310--1318. PMLR, 2013.

\bibitem{passalis2018unsupervised}
Nikolaos Passalis and Anastasios Tefas.
\newblock Unsupervised knowledge transfer using similarity embeddings.
\newblock {\em IEEE transactions on neural networks and learning systems},
  30(3):946--950, 2018.

\bibitem{NEURIPS2019_9015}
Adam Paszke, Sam Gross, Francisco Massa, Adam Lerer, James Bradbury, Gregory
  Chanan, Trevor Killeen, Zeming Lin, Natalia Gimelshein, Luca Antiga, Alban
  Desmaison, Andreas Kopf, Edward Yang, Zachary DeVito, Martin Raison, Alykhan
  Tejani, Sasank Chilamkurthy, Benoit Steiner, Lu~Fang, Junjie Bai, and Soumith
  Chintala.
\newblock Pytorch: An imperative style, high-performance deep learning library.
\newblock In H.~Wallach, H.~Larochelle, A.~Beygelzimer, F.~d\textquotesingle
  Alch\'{e}-Buc, E.~Fox, and R.~Garnett, editors, {\em Advances in Neural
  Information Processing Systems 32}, pages 8024--8035. Curran Associates,
  Inc., 2019.

\bibitem{reed2014training}
Scott Reed, Honglak Lee, Dragomir Anguelov, Christian Szegedy, Dumitru Erhan,
  and Andrew Rabinovich.
\newblock Training deep neural networks on noisy labels with bootstrapping.
\newblock {\em arXiv preprint arXiv:1412.6596}, 2014.

\bibitem{russakovsky2015imagenet}
Olga Russakovsky, Jia Deng, Hao Su, Jonathan Krause, Sanjeev Satheesh, Sean Ma,
  Zhiheng Huang, Andrej Karpathy, Aditya Khosla, Michael Bernstein, et~al.
\newblock Imagenet large scale visual recognition challenge.
\newblock {\em International journal of computer vision}, 115(3):211--252,
  2015.

\bibitem{smith2020generalization}
Samuel Smith, Erich Elsen, and Soham De.
\newblock On the generalization benefit of noise in stochastic gradient
  descent.
\newblock In {\em International Conference on Machine Learning}, pages
  9058--9067. PMLR, 2020.

\bibitem{szegedy2016rethinking}
Christian Szegedy, Vincent Vanhoucke, Sergey Ioffe, Jon Shlens, and Zbigniew
  Wojna.
\newblock Rethinking the inception architecture for computer vision.
\newblock In {\em Proceedings of the IEEE conference on computer vision and
  pattern recognition}, pages 2818--2826, 2016.

\bibitem{wang2018cosface}
Hao Wang, Yitong Wang, Zheng Zhou, Xing Ji, Dihong Gong, Jingchao Zhou, Zhifeng
  Li, and Wei Liu.
\newblock Cosface: Large margin cosine loss for deep face recognition.
\newblock In {\em Proceedings of the IEEE conference on computer vision and
  pattern recognition}, pages 5265--5274, 2018.

\bibitem{wang2021memory}
Jiyue Wang, Pei Zhang, and Yanxiong Li.
\newblock Memory-replay knowledge distillation.
\newblock {\em Sensors}, 21(8):2792, 2021.

\bibitem{wang2020real}
Ning Wang, Wengang Zhou, Yibing Song, Chao Ma, and Houqiang Li.
\newblock Real-time correlation tracking via joint model compression and
  transfer.
\newblock {\em IEEE Transactions on Image Processing}, 29:6123--6135, 2020.

\bibitem{rw2019timm}
Ross Wightman.
\newblock Pytorch image models.
\newblock \url{https://github.com/rwightman/pytorch-image-models}, 2019.

\bibitem{wu2018improving}
Zhirong Wu, Alexei~A Efros, and Stella~X Yu.
\newblock Improving generalization via scalable neighborhood component
  analysis.
\newblock In {\em Proceedings of the European Conference on Computer Vision
  (ECCV)}, pages 685--701, 2018.

\bibitem{xie2016disturblabel}
Lingxi Xie, Jingdong Wang, Zhen Wei, Meng Wang, and Qi~Tian.
\newblock Disturblabel: Regularizing cnn on the loss layer.
\newblock In {\em Proceedings of the IEEE Conference on Computer Vision and
  Pattern Recognition}, pages 4753--4762, 2016.

\bibitem{xu2019data}
Ting-Bing Xu and Cheng-Lin Liu.
\newblock Data-distortion guided self-distillation for deep neural networks.
\newblock In {\em Proceedings of the AAAI Conference on Artificial
  Intelligence}, volume~33, pages 5565--5572, 2019.

\bibitem{yao2018deep}
Jiangchao Yao, Jiajie Wang, Ivor~W Tsang, Ya~Zhang, Jun Sun, Chengqi Zhang, and
  Rui Zhang.
\newblock Deep learning from noisy image labels with quality embedding.
\newblock {\em IEEE Transactions on Image Processing}, 28(4):1909--1922, 2018.

\bibitem{yuan2020revisiting}
Li~Yuan, Francis~EH Tay, Guilin Li, Tao Wang, and Jiashi Feng.
\newblock Revisiting knowledge distillation via label smoothing regularization.
\newblock In {\em Proceedings of the IEEE/CVF Conference on Computer Vision and
  Pattern Recognition}, pages 3903--3911, 2020.

\bibitem{yun2020regularizing}
Sukmin Yun, Jongjin Park, Kimin Lee, and Jinwoo Shin.
\newblock Regularizing class-wise predictions via self-knowledge distillation.
\newblock In {\em Proceedings of the IEEE/CVF conference on computer vision and
  pattern recognition}, pages 13876--13885, 2020.

\bibitem{zhai2018classification}
Andrew Zhai and Hao-Yu Wu.
\newblock Classification is a strong baseline for deep metric learning.
\newblock {\em arXiv preprint arXiv:1811.12649}, 2018.

\bibitem{zhang2019gradient}
Jingzhao Zhang, Tianxing He, Suvrit Sra, and Ali Jadbabaie.
\newblock Why gradient clipping accelerates training: A theoretical
  justification for adaptivity.
\newblock {\em arXiv preprint arXiv:1905.11881}, 2019.

\bibitem{zhang2019your}
Linfeng Zhang, Jiebo Song, Anni Gao, Jingwei Chen, Chenglong Bao, and Kaisheng
  Ma.
\newblock Be your own teacher: Improve the performance of convolutional neural
  networks via self distillation.
\newblock In {\em Proceedings of the IEEE/CVF International Conference on
  Computer Vision}, pages 3713--3722, 2019.

\bibitem{zhang2018deep}
Ying Zhang, Tao Xiang, Timothy~M Hospedales, and Huchuan Lu.
\newblock Deep mutual learning.
\newblock In {\em Proceedings of the IEEE Conference on Computer Vision and
  Pattern Recognition}, pages 4320--4328, 2018.

\end{thebibliography}

\clearpage

\section{Supplementary section}
\label{sec:Supp}

\subsection{Sample Code}

In PyTorch, the available method to modify gradients in the backward pass is using the
\codeword{register\_hook} function which is called in the \codeword{forward} function.

\begin{figure}[ht] \hrule 
\begin{lstlisting}[language=Python]
# Modification of gradients take place
# inside the forward function
def forward(self, x, labels):
    # x: Input tensor with shape [N, C, H, W]
    # labels: One hot ground truth with shape [N, C]
    logits = self.layers(x)

    def powerGradTransform(grad):
        alpha = 0.3
        # Recover the predicted probabilites
        # generated in the forward pass
        grad_temp = grad / grad.shape[0]
        pred = grad_temp + labels
        # Transform the predicted probabilities
        pred = pred**alpha
        pred /= pred.sum(-1, keepdim=True)
        # Modify the gradient
        modified_grad = pred - labels
        modified_grad = modified_grad * grad.shape[0]
        return modified_grad

    if self.training:
        logits.register_hook(powerGradTransform)
    return logits
\end{lstlisting} \hrule
\vspace{0.1cm} \caption{Python code of PowerGrad Transform based on PyTorch. }
\label{fig:code}
\end{figure}

\clearpage

\subsection{Grid search for different values of $\alpha$}

\begin{figure}[ht]
\centering
\includegraphics[width=0.5\columnwidth]{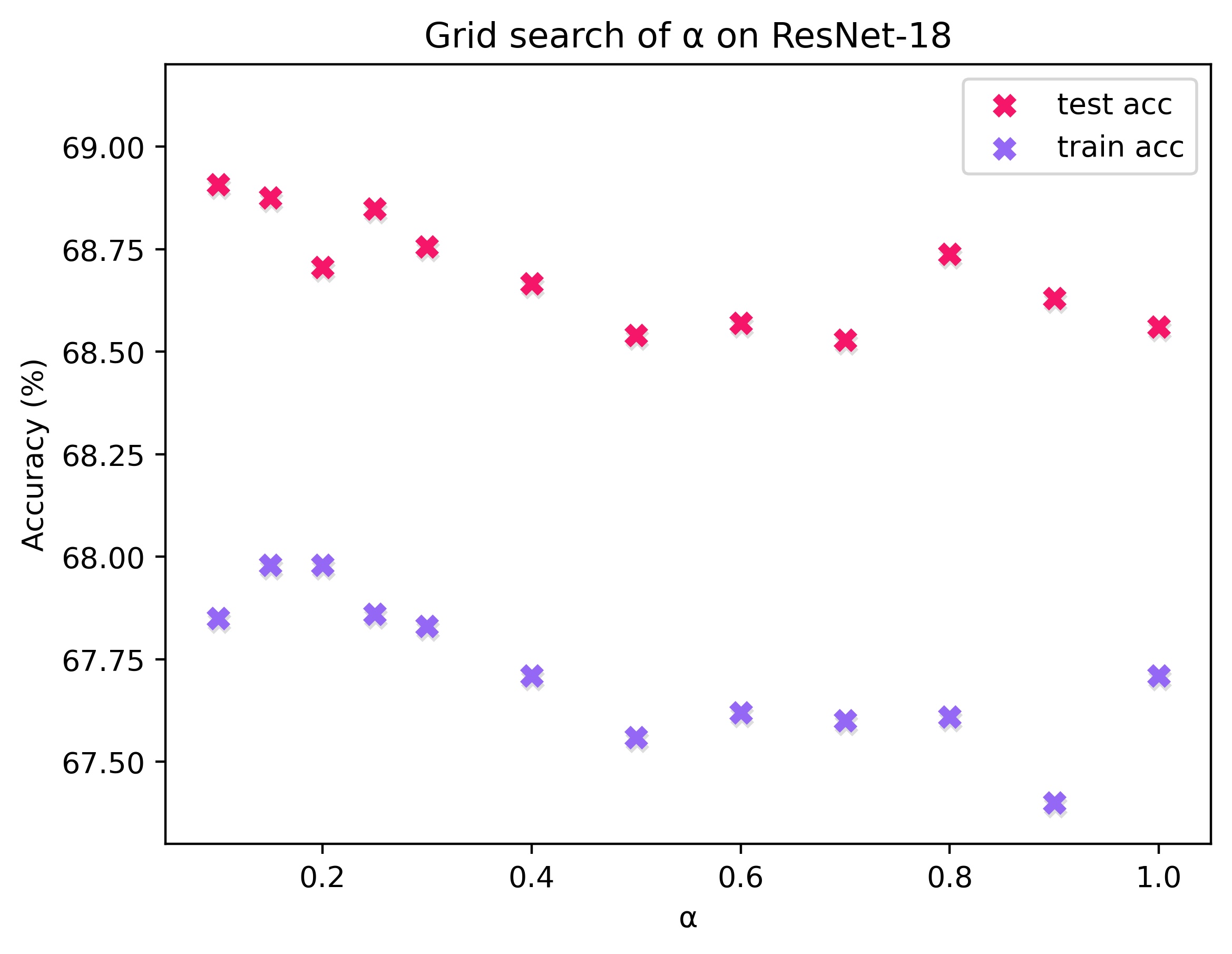}
\caption{ Grid search for various values of $\alpha$ for ResNet-18 architecture with the
ImageNet dataset. Each datapoint is collected over a 50 epoch budget, with other
hyperparameters kept the same as mentioned in section \ref{sec:bn}. }
\label{fig:grid_searcha}
\end{figure}

\begin{figure}[ht]
\centering
\includegraphics[width=0.5\columnwidth]{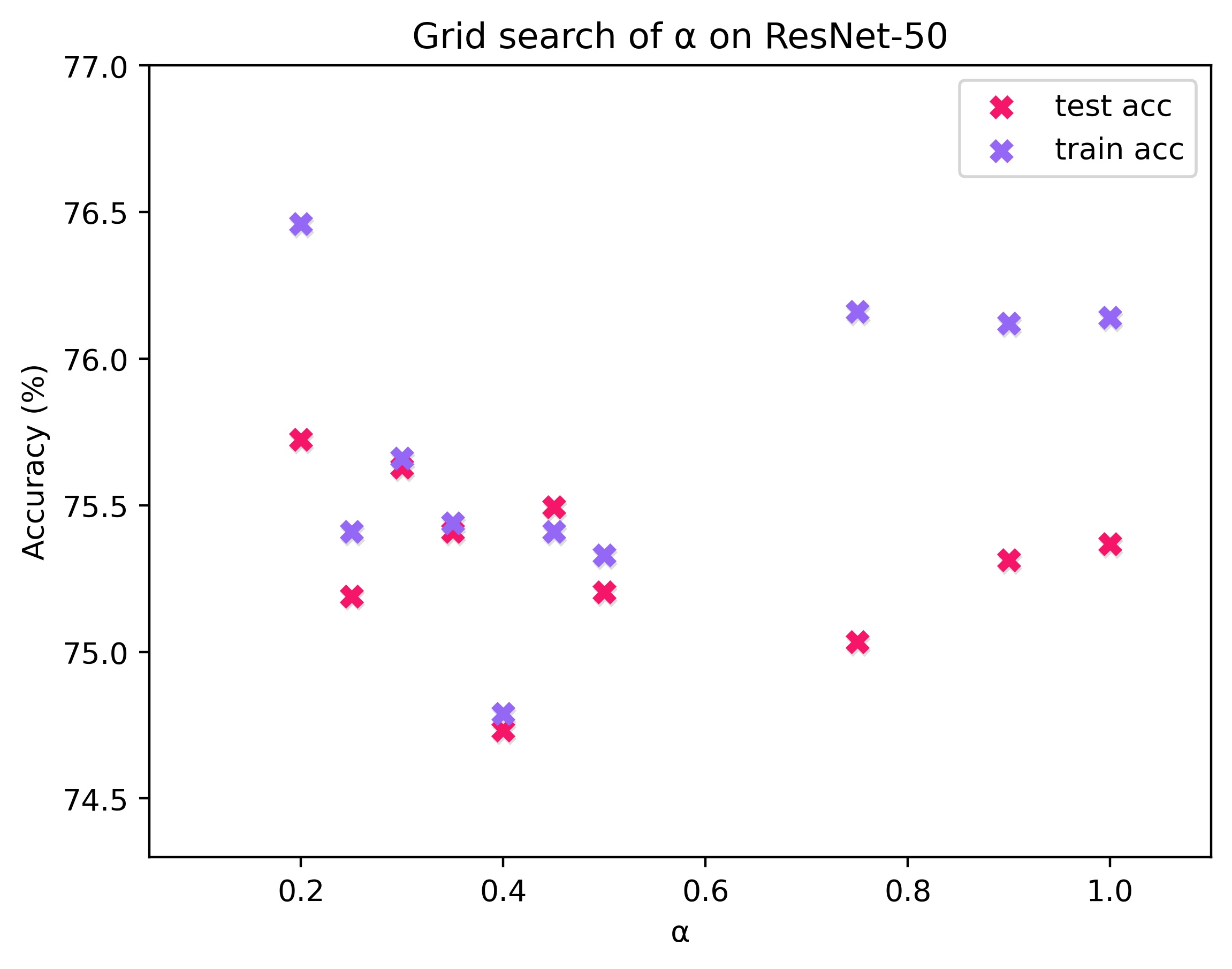}
\caption{ Grid search for various values of $\alpha$ for ResNet-50 architecture with the
ImageNet dataset. Each datapoint is collected over a 50 epoch budget, with other
hyperparameters kept the same as mentioned in section \ref{sec:bn}. }
\label{fig:grid_searchb}
\end{figure}

\clearpage

\begin{figure}[ht]
\centering
\includegraphics[width=0.5\columnwidth]{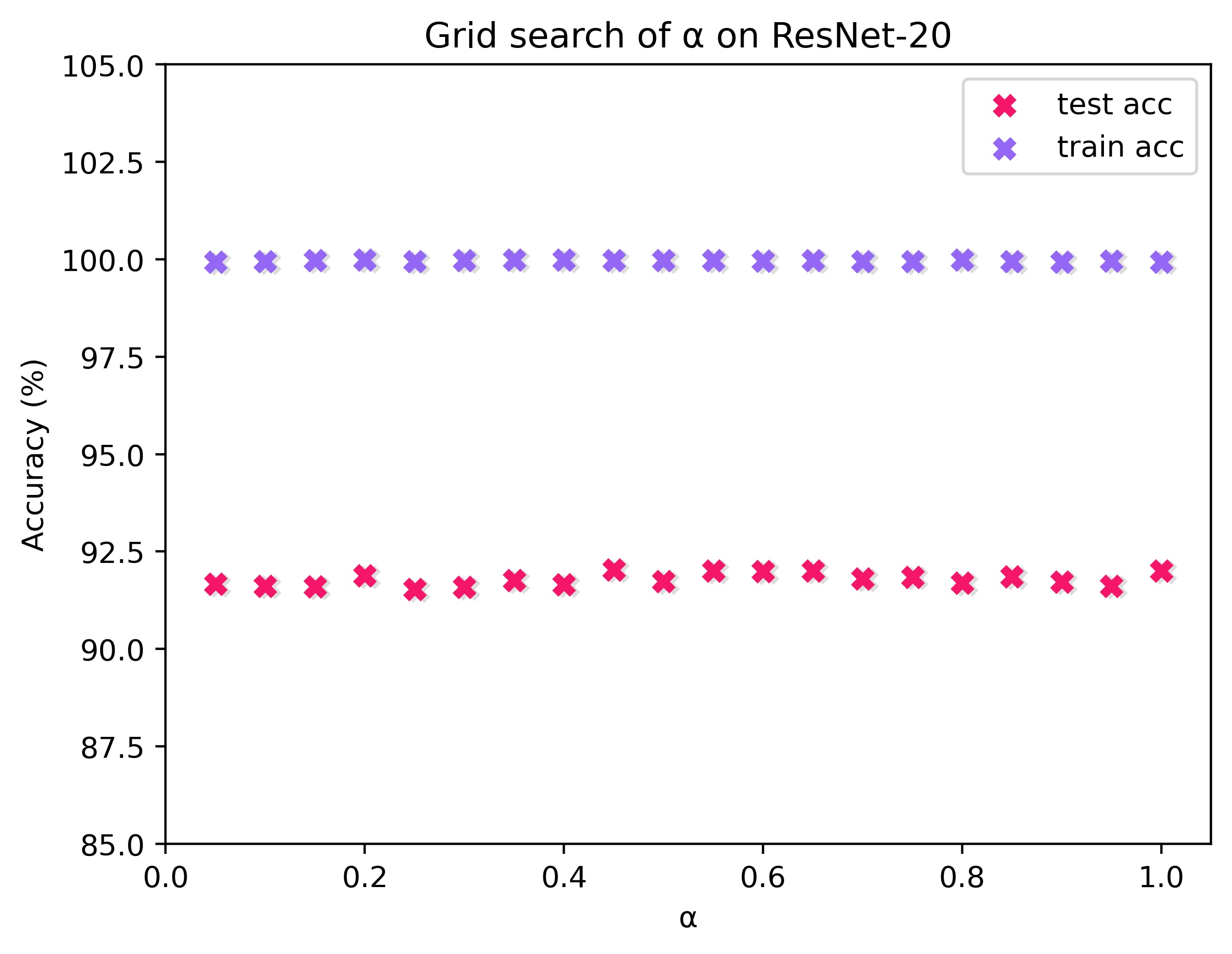}
\caption{\\ResNet-20 CIFAR-10}
\caption{ Grid search for various values of $\alpha$ for ResNet-20 architecture with the
CIFAR-10 dataset. Each datapoint is collected over a 360 epoch budget. }
\label{fig:grid_searchc}
\end{figure}

There is no effect of PGT on small datasets such as CIFAR-10. In these scenarios,
networks almost always achieve $100\%$ training accuracy, leaving no room for PGT to
assist.

\clearpage

\subsection{Norm plots of weights (baseline (low batch size)) of non-BN ResNet-18}
\begin{figure}[ht] \centering \includegraphics[width=0.55\textwidth]{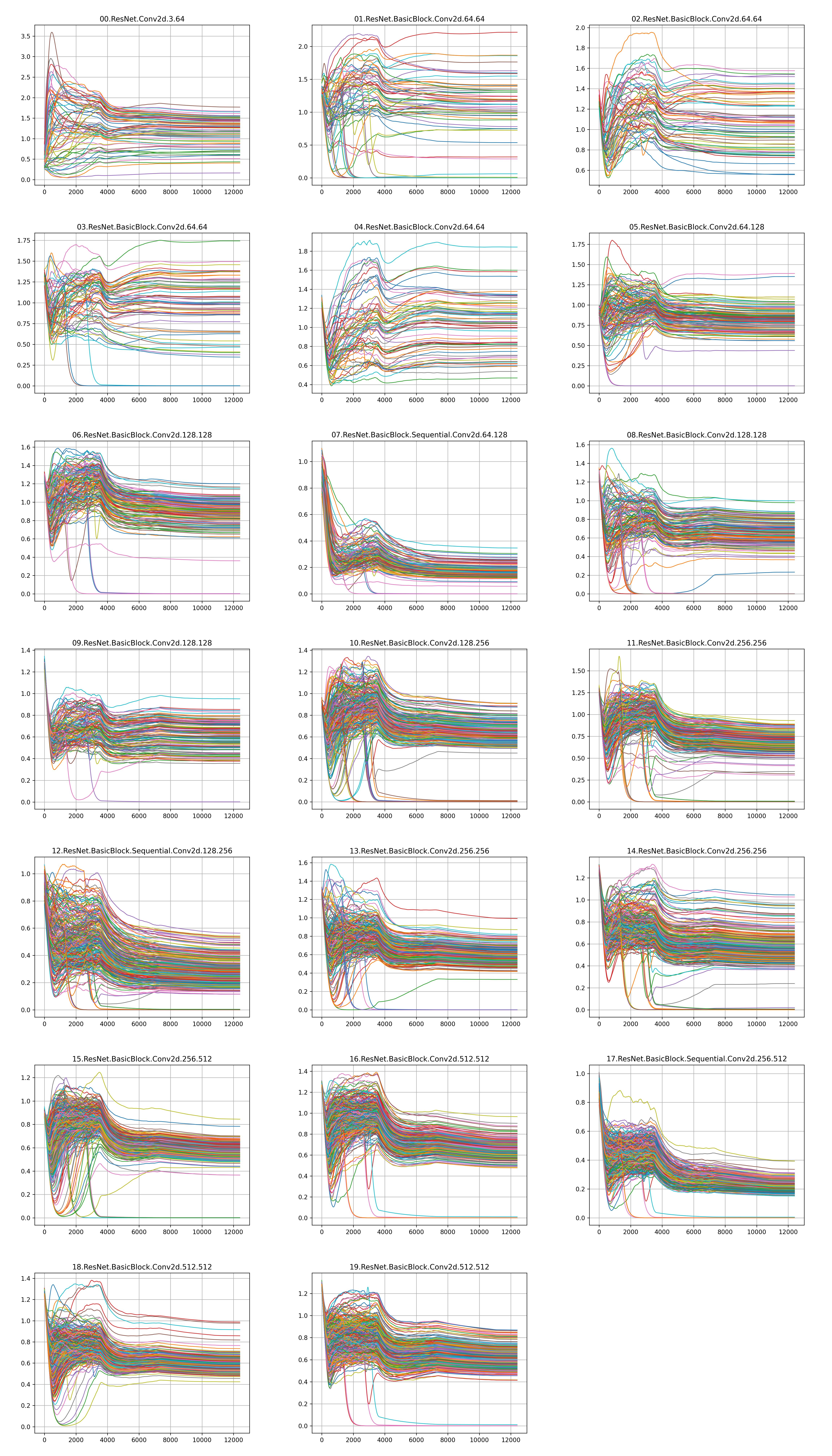}
\caption{ Method: baseline, batch size = 256. This is a plot depicting the
iteration-wise evolution of the norm of each filter in each layer. Each subplot
corresponds to a layer of the unnormalized ResNet-18. Above each subplot, the layer
indices are shown. Each filter in each subplot is shown with a separate and randomly
determined colour. } \end{figure}

\clearpage

\subsection{Norm plots of weights (baseline (high batch size)) of non-BN ResNet-18}
\begin{figure}[ht] \centering
\includegraphics[width=0.55\textwidth]{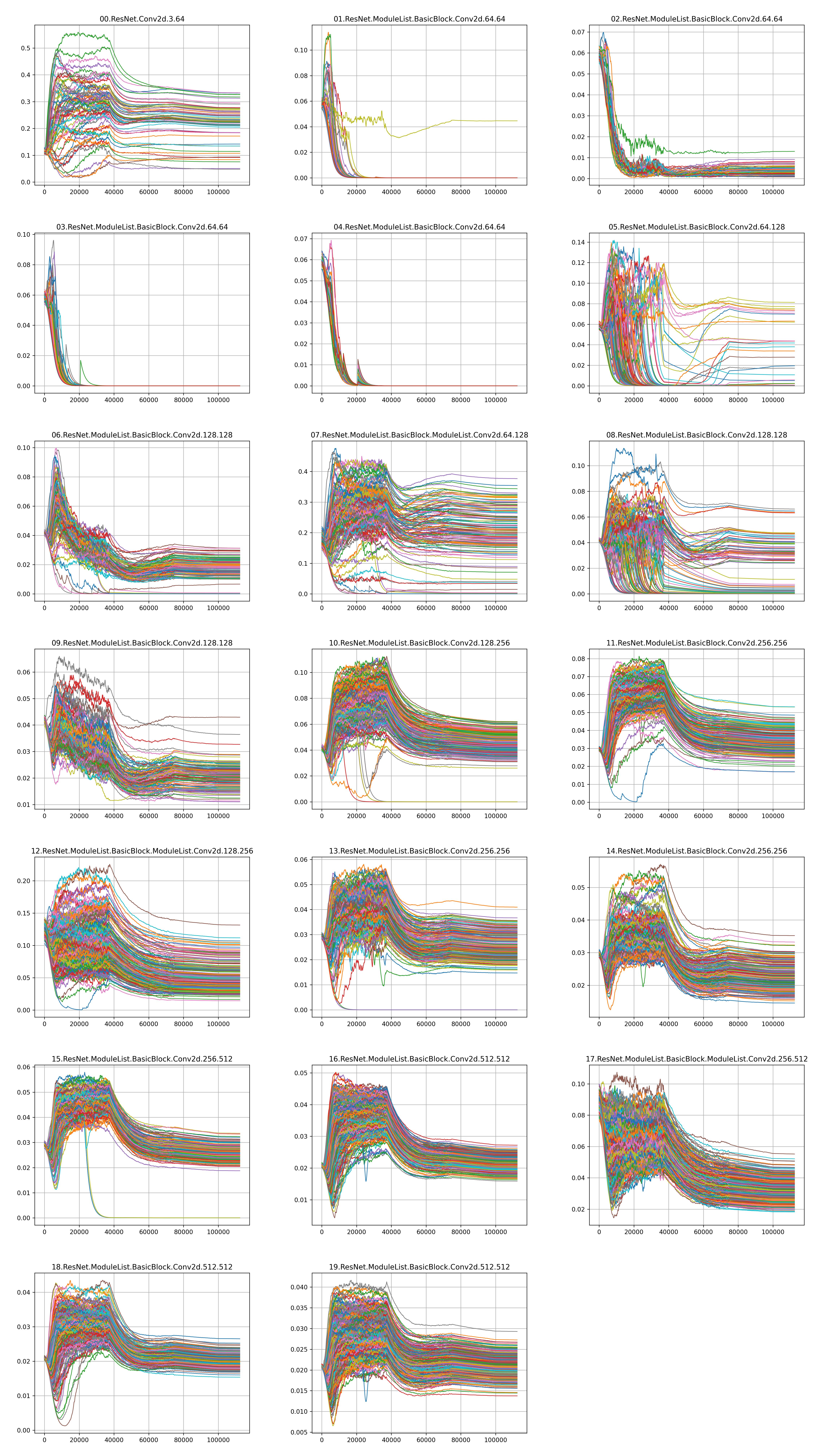} \caption{ Method:
baseline (high batch size), batch size = 1024. This is a plot depicting the
iteration-wise evolution of the norm of each filter in each layer. Each subplot
corresponds to a layer of the unnormalized ResNet-18. Above each subplot, the layer
indices are shown. Each filter in each subplot is shown with a separate and randomly
determined colour. } \end{figure}

\clearpage

\subsection{Norm plots of weights (PGT) of non-BN ResNet-18}
\begin{figure}[ht] \centering \includegraphics[width=0.55\textwidth]{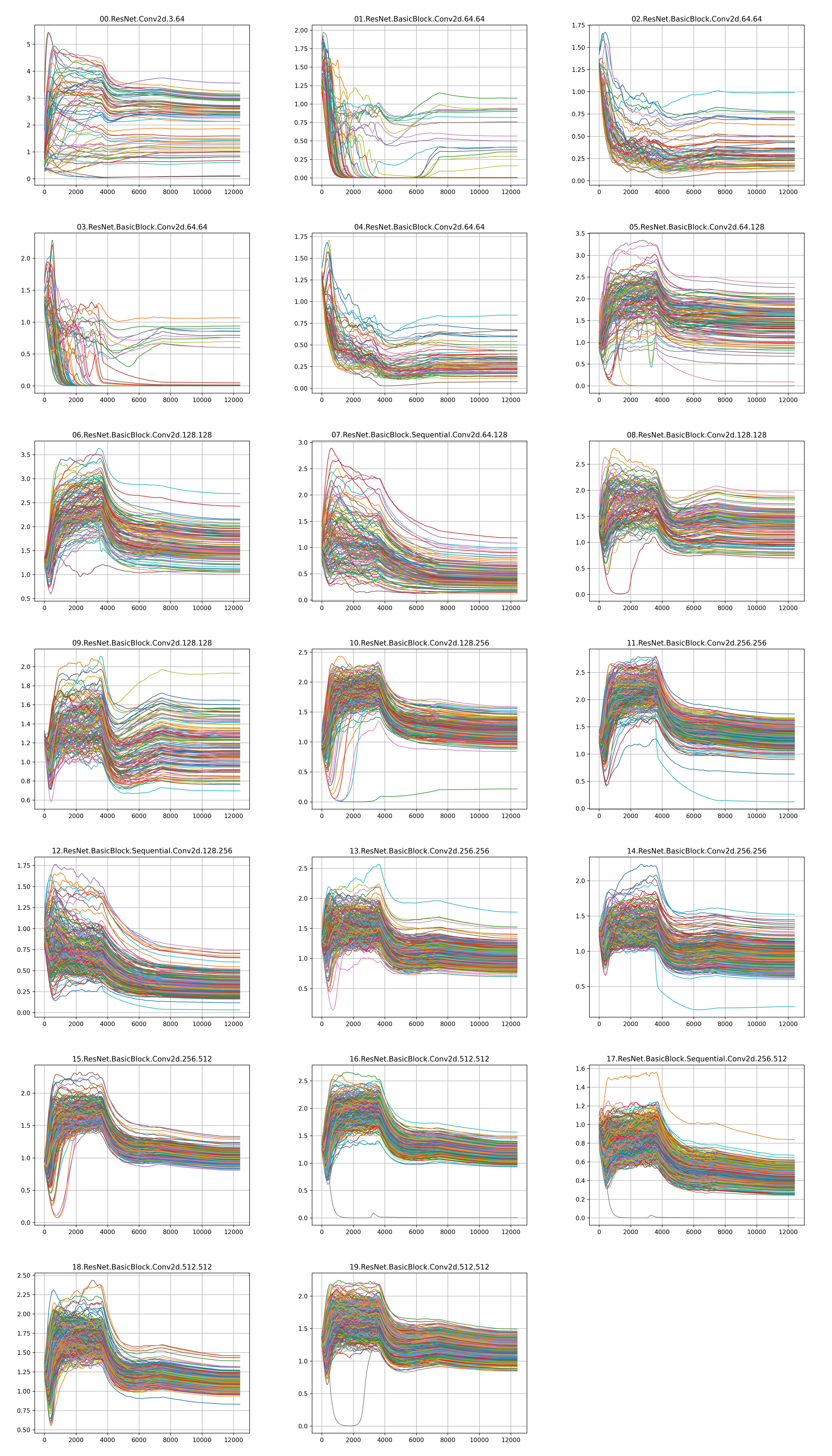}
\caption{ Method: PGT, batch size = 256. This is a plot depicting the iteration-wise
evolution of the norm of each filter in each layer. Each subplot corresponds to a layer
of the unnormalized ResNet-18. Above each subplot, the layer indices are shown. Each
filter in each subplot is shown with a separate and randomly determined colour. }
\end{figure}

\clearpage

\subsection{Norm plots of weights (AGC + PGT) of non-BN ResNet-18}
\begin{figure}[ht] \centering \includegraphics[width=0.55\textwidth]{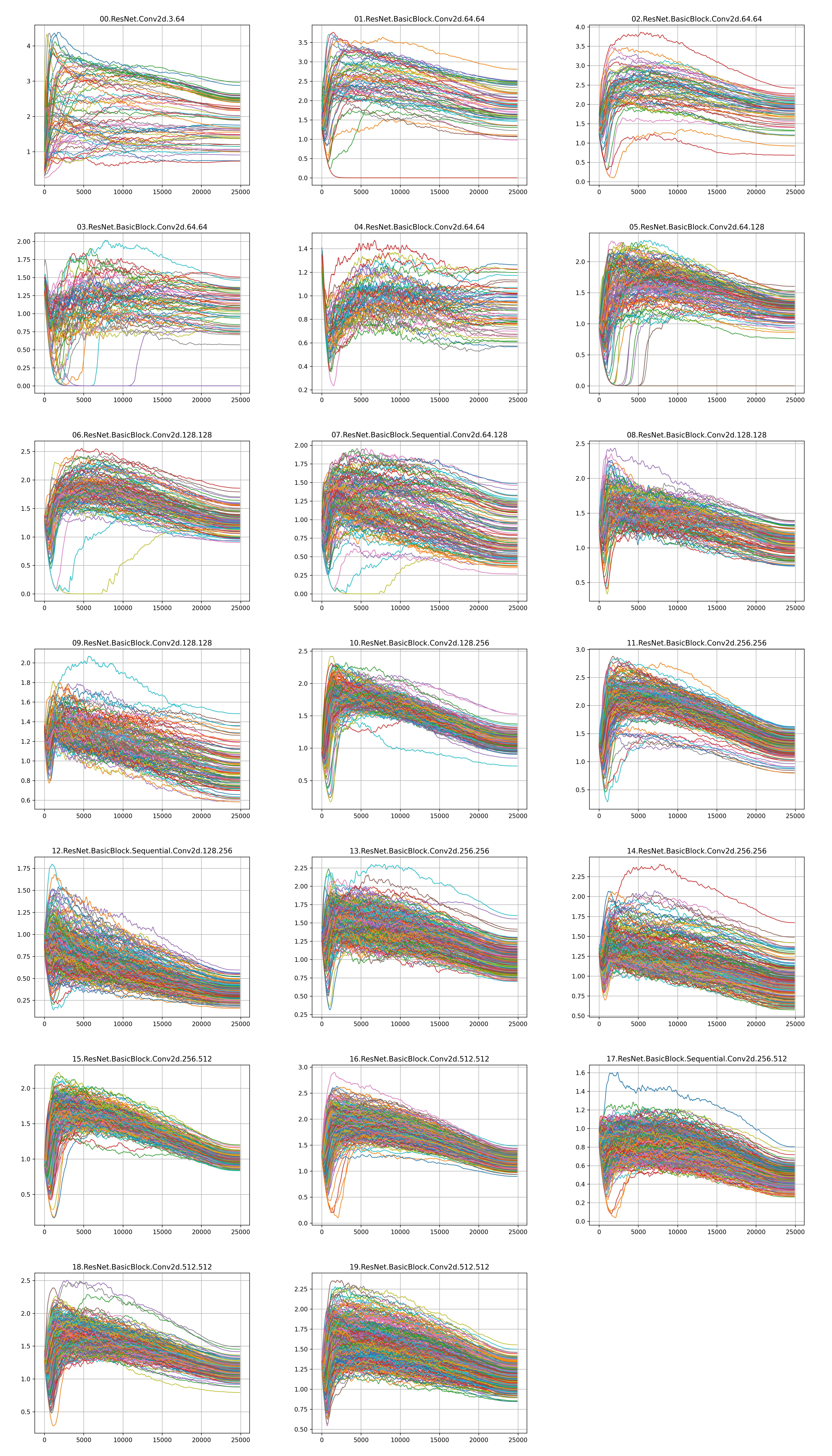}
\caption{ Method: AGC + PGT, batch size = 256. This is a plot depicting the
iteration-wise evolution of the norm of each filter in each layer. Each subplot
corresponds to a layer of the unnormalized ResNet-18. Above each subplot, the layer
indices are shown. Each filter in each subplot is shown with a separate and randomly
determined colour. } \end{figure}

\clearpage

\subsection{ResNet-18 Layer Index Diagram}

\begin{figure}[ht] \centering \includegraphics[width=0.96\columnwidth]{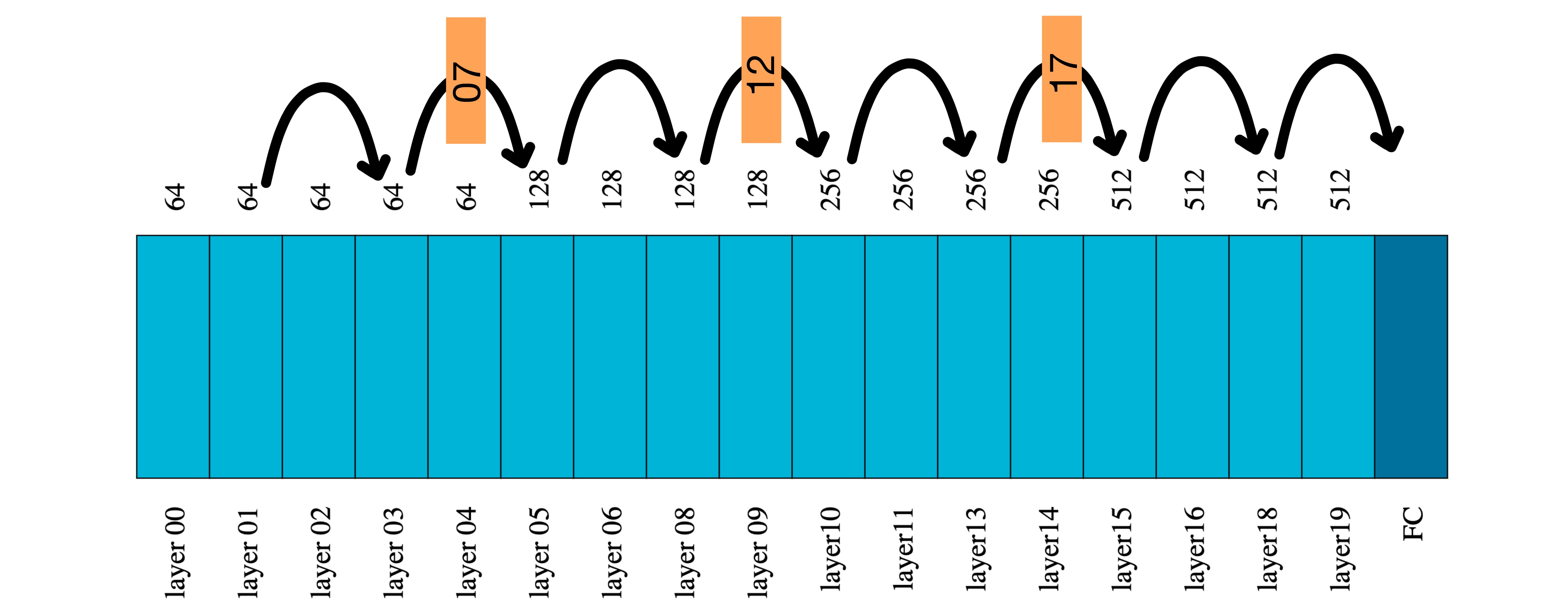}
\caption{ ResNet-18 architecture with layer indices of convolutional layers and the
number of filters in each layer. } \label{fig:resnet18} \end{figure}

For the layer indices used in figures Fig. \ref{fig:norm_plots}, \ref{fig:high_bs},
\ref{fig:agc_pgt}, we provide a block diagram depicting the different convolutional
layers of ResNet-18 with corresponding layer indices in Fig. \ref{fig:resnet18}. Each
layer represents a convolutional layer, with the layer index and the number of filters
denoted alongside it. Downsampling convolutional layers in skip connections are drawn
and their layer indices are shown on the skip connection paths.

ResNet-18 is composed of $20$ convolutional layers and one fully connected (FC) layer.
At the input, it has a single convolutional layer with filters of size $7\times 7$.
Three convolutional layers are placed in the downsampling skip connections. The
remaining convolutional layers have several $3\times 3$ filters that are grouped
together to create a block. Each basic block is connected by a skip connection. At the
end of all convolutional layers, a global average pooling layer downsamples the features
to a fixed-size feature vector and transfers it to the FC layer, which fits the features
using regression and outputs logits. The logits $z_i$ produced by the FC layer are fed
into a softmax layer, which transforms them to predicted probabilities $p_i$ using the
equation $p_i=\frac {e^{z_i}}{\sum _j e^{z_j}}$.

\end{document}